%% file: bourdin.tex
\def\papertitle{Empirical Results for Adjusting Truncated Backpropagation Through Time while Training Neural Audio Effects}
\def\paperauthorA{Yann Bourdin}
\def\paperauthorB{Pierrick Legrand}
\def\paperauthorC{Fanny Roche}
\documentclass[twoside,a4paper]{article}
\usepackage{etoolbox}
\usepackage{dafx_25}
\usepackage{amsmath,amssymb,amsfonts,amsthm}
\usepackage{euscript}
\usepackage[T1]{fontenc}
\usepackage[utf8]{inputenc}
\usepackage{ifpdf}
\usepackage[english]{babel}

\usepackage{booktabs}
\usepackage{color}
\usepackage{url}

\usepackage{stmaryrd}
\usepackage{multicol}
\usepackage{adjustbox}
\usepackage{subcaption}
\usepackage[table]{xcolor}
\usepackage{tabularx}
\renewcommand\tabularxcolumn[1]{m{#1}}% for vertical centering text in X column % étrangement ça affecte aussi la hauteur des cellules, il n'y a plus besoin de arraystretch
\usepackage{multirow}
\usepackage{wrapfig}
\usepackage{afterpage}

\input glyphtounicode
\ninept

\newcounter{numauth}
\newcounter{listcnt}
\newcommand\authcnt[1]{\ifdefined#1 \stepcounter{numauth} \fi}

\newcommand\addauth[1]{
\ifdefined#1 
\stepcounter{listcnt}
\ifnum \value{listcnt}<\value{numauth}
\appto\authorslist{, #1}
\else
\appto\authorslist{~and~#1}
\fi
\fi}
\authcnt{\paperauthorB}
\authcnt{\paperauthorC}
\authcnt{\paperauthorD}
\authcnt{\paperauthorE}
\authcnt{\paperauthorF}
\authcnt{\paperauthorG}
\authcnt{\paperauthorH}
\authcnt{\paperauthorI}
\authcnt{\paperauthorJ}
\def\authorslist{\paperauthorA}
\addauth{\paperauthorB}
\addauth{\paperauthorC}
\addauth{\paperauthorD}
\addauth{\paperauthorE}
\addauth{\paperauthorF}
\addauth{\paperauthorG}
\addauth{\paperauthorH}
\addauth{\paperauthorI}
\addauth{\paperauthorJ}
\usepackage{times}
\newif\ifpdf
\ifx\pdfoutput\relax
\else
   \ifcase\pdfoutput
      \pdffalse
   \else
      \pdftrue
   \fi
\fi

  \usepackage[pdftex,
    pdftitle={\papertitle},
    pdfauthor={\authorslist},
    pdfsubject={Proceedings of the 28th International Conference on Digital Audio Effects (DAFx25)},
    colorlinks=false, % links are activated as color boxes instead of color text
    bookmarksnumbered, % use section numbers with bookmarks
    pdfstartview=XYZ % start with zoom=100% instead of full screen; especially useful if working with a big screen :-)
  ]{hyperref}
\usepackage[hypcap=true]{caption}
\title{\papertitle}

\threeaffiliations{
\paperauthorA \,\thanks{\vspace{-3mm}}}
{Arturia \\ Montbonnot-Saint-Martin, France \\[1mm] Astral Inria Team, Inria Bordeaux \\ Talence, France \\
{\tt \href{yann.bourdin@arturia.com}{yann.bourdin@arturia.com}}
}
{\paperauthorB}
{IMS, UMR CNRS 5218 \\ ENSC, Bordeaux INP \\ Astral Inria Team, Inria Bordeaux \\ Talence, France \\
{\tt \href{mailto:pierrick.legrand@ensc.fr}{pierrick.legrand@ensc.fr}}
}
{\paperauthorC}
{Arturia \\ Montbonnot-Saint-Martin, France \\
{\tt \href{fanny.roche@arturia.com}{fanny.roche@arturia.com}}
}
\begin{document}
% more pdf-tex settings:
\ifpdf % used graphic file format for pdflatex
  \DeclareGraphicsExtensions{.png,.jpg,.pdf}
\else  % used graphic file format for latex
  \DeclareGraphicsExtensions{.eps}
\fi

%\makeatletter
%\pdfbookmark[0]{\@pdftitle}{title}
%\makeatother

\maketitle

\begin{abstract}
This paper investigates the optimization of Truncated Backpropagation Through Time (TBPTT) for training neural networks in digital audio effect modeling, with a focus on dynamic range compression. The study evaluates key TBPTT hyperparameters -- sequence number, batch size, and sequence length -- and their influence on model performance. Using a convolutional-recurrent architecture, we conduct extensive experiments across datasets with and without conditioning by user controls. Results demonstrate that carefully tuning these parameters enhances model accuracy and training stability, while also reducing computational demands. Objective evaluations confirm improved performance with optimized settings, while subjective listening tests indicate that the revised TBPTT configuration maintains high perceptual quality.
\end{abstract}

\section{Introduction}

Audio effects are vital in shaping music, influencing timbre, dynamics, and spatial traits in both production and performance. Initially developed with analog circuitry, digital emulation is now important for its portability, flexibility, and lower cost. However, capturing analog devices' unique sonic traits digitally is challenging due to their nonlinear and time-dependent behaviors. Traditional modeling methods include white-box, gray-box, and black-box approaches \cite{comunita2025differentiable}. White-box modeling uses precise mathematical descriptions of components, offering accuracy but requiring deep knowledge and high computational cost. Black-box methods replicate effects based on input-output data without detailed internal understanding. Gray-box models merge both approaches, integrating partial system knowledge with data-driven techniques.
Deep learning has emerged as a promising black-box method for audio effect modeling, enabling neural networks to learn complex signal transformations only from input/output recordings, removing the need for handcrafted equations, extensive tuning, and in-depth hardware knowledge. Despite its potential, challenges remain in modeling effects with parameters and long time dependencies. 

This work builds on \cite{bourdinTacklingLongRangeDependencies2024}, focusing on Dynamic Range Compression (DRC).
A compressor modifies an audio signal's dynamics by applying time-varying gain reduction based on the input or sidechain signal's level. The nonlinear, time-dependent, time-invariant\footnote{The compression parameters do not vary over time, but the gain reduction depends on the past of the input or sidechain signal.} 
nature and parameter conditioning of compressors make them relevant for this research. A compressor's behavior is mainly governed by four parameters: Threshold, defining the level above which gain reduction is applied; Ratio, determining attenuation degree; Attack time, controlling how quickly compression starts; and Release time, setting how gradually gain reduction is lifted when the signal drops below the threshold.
Neural modeling of DRC has been explored using various architectures, including auto-encoders in the spectral domain \cite{hawleySignalTrainProfilingAudio2019}, convolutional networks \cite{steinmetzEfficientNeuralNetworks2022}, recurrent networks \cite{simionatoFULLYCONDITIONEDLOWLATENCY2023}, and convolutional-recurrent networks \cite{ramirezDeepLearningAudio2021, comunitaModellingBlackBoxAudio2023, bourdinTacklingLongRangeDependencies2024}. However, large attack and release times introduce long time dependencies, complicating modeling due to the need for handling extensive audio sequence lengths.
In \cite{bourdinTacklingLongRangeDependencies2024}, the State Prediction Network (SPN) was introduced to efficiently train recurrent networks with long time dependencies. This convolutional neural network predicts the model's initial states, replacing the processing needed to warm up these states, thus reducing training times. However, state prediction can increase error in inference compared to training due to exposure bias \cite{peussaExposureBiasState2021}, i.e., exposure to ground truth during training. This discrepancy motivates reducing reliance on the SPN through an approach derived from Truncated Backpropagation Through Time (TBPTT) \cite{elman1990finding}, to be introduced in Section \ref{methods}. While commonly used, TBPTT parameters are typically set empirically. In Section \ref{experiments}, our main contribution is to demonstrate, by training numerous models, the critical importance of properly setting these parameters alongside the batch size. Our results are validated by an objective evaluation and a subjective listening test. 

\section{Methods}
\label{methods}

\subsection{Model Architecture}
\label{model}

\begin{figure}[t]
\begin{center}
\includegraphics[width=0.95\columnwidth]{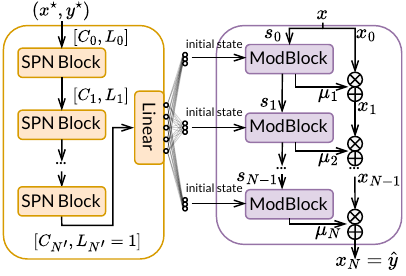}
\caption{State prediction network (SPN) and SPTMod}
\label{fig:spn_sptmod}
\end{center}
\end{figure}

\begin{figure}[t]
\centering
\includegraphics[width=0.4\linewidth, trim={0 -4em 0 0}]{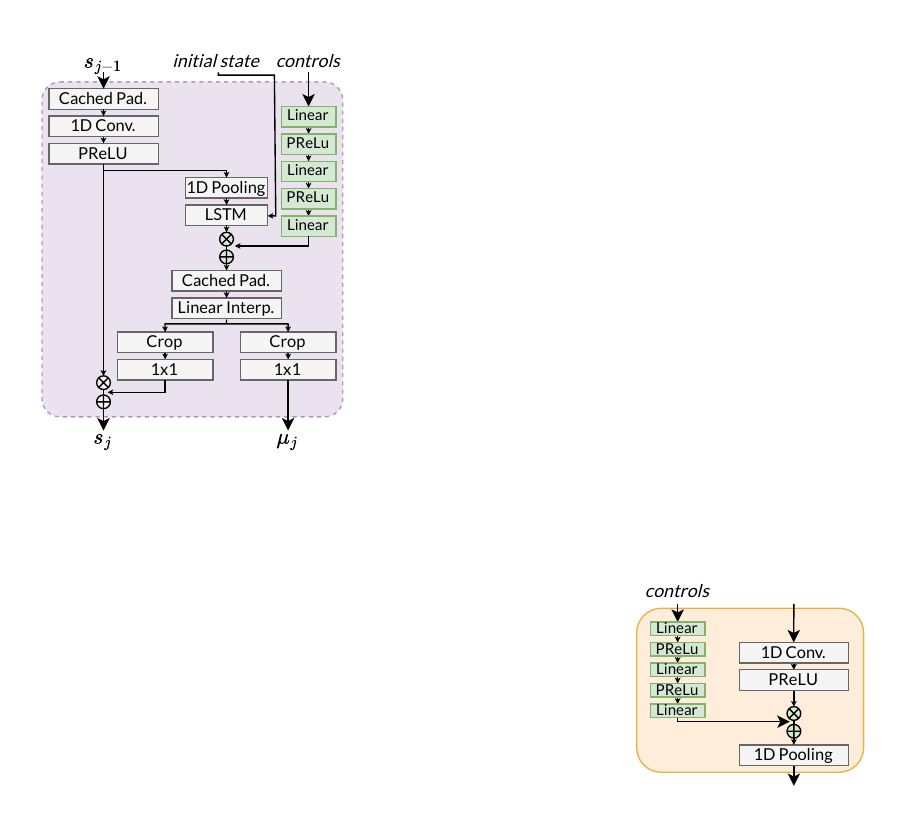}
\hfill
\includegraphics[width=0.58\linewidth]{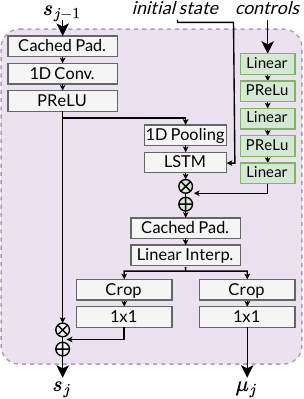}
\caption{State Prediction Block (SPN Block, on the left side), and Modulation Block (ModBlock, on the right side).}
\vspace{-1em}
\label{fig:blocks}
\end{figure}

The architecture we focus on, SPTMod (Series-Parallel Temporal Modulation), is a convolutional-recurrent network initially designed for DRC, as introduced in \cite{bourdinTacklingLongRangeDependencies2024}. It is similar to the GCNTFiLM model \cite{comunitaModellingBlackBoxAudio2023} but is lighter as it involves fewer nonlinear operations.
As shown in Figure \ref{fig:spn_sptmod}, SPTMod's backbone includes two paths: the modulation path and the audio path. The modulation path comprises blocks called ModBlock (typically 3 blocks, see Section \ref{hyperparameters}) that calculate modulations for the audio path tensors via Feature-wise Linear Modulation (FiLM) operations \cite{perezFiLMVisualReasoning2017}. These operations scale and shift tensor channels using modulation tensors. Since the target effect is a compressor, to ensure the final audio output is not degraded, the audio path performs only amplitude modulation through Temporal FiLM (TFiLM) operations. Though the architecture avoids convolutional layers in the audio path, adding them could enable dynamic convolution for modeling other audio effects \cite{comunitaModellingBlackBoxAudio2023, chen2020dynamic}.
Each modulation block, detailed in Figure \ref{fig:blocks}, starts with a 1D convolutional layer followed by a Parametric Rectified Linear Unit (PReLU) activation. The output is processed by a TFiLM sub-block, which includes a pooling layer for downsampling, followed by a Long Short-Term Memory (LSTM) layer. This is complemented by a FiLM operation, transforming user controls via a two-layer neural network. A linear interpolation layer for upsampling concludes TFiLM to compensate for pooling. Two 1D convolutional layers with a kernel size of 1 (`1x1') adjust channel count before FiLM operations, yielding tensors \( s_j \) and \( x_j \). Note that the LSTM layer, being recurrent, has an internal state whose initial value should be seen as an input to the modulation block. Cropping and cached padding layers are detailed in Section \ref{temporal}.

Proper initialization of the LSTM layers' initial states is crucial due to the effect's time dependency. We use the State Prediction Network (SPN), as introduced in \cite{bourdinTacklingLongRangeDependencies2024} and shown in Figure \ref{fig:spn_sptmod}, to address this. Typically, states are initialized with a warm-up method, processing a number of samples before gradients are registered. However, for effects with long time dependencies, even longer than the sequence length used for training, the warm-up length can become computationally expensive. The SPN suits these cases by efficiently reducing tensor size exponentially.
The SPN receives inputs and reference outputs of lengths close to the effect's time dependency and predicts the state values the model would generate processing this input. Designed like a convolutional network performing classification, it consists of blocks, each with a 1D convolution, a PReLU activation, and a pooling layer, as seen in Figure \ref{fig:blocks}. The SPN is also conditioned by user controls through the same FiLM method as used in the modulation blocks. After each pooling layer, the intermediary tensor's temporal length is divided by a factor, shrinking it exponentially with block depth. The final SPN block reduces the length to one, resulting in a batched vector. A final linear layer transforms this vector to match the sum of state sizes, allowing initial states of the model to be derived from slices of this vector.

An implementation of SPTMod is provided at the accompanying repository.\footnote{\url{https://github.com/ybourdin/sptmod}}

\subsection{Loss}
\label{loss}

The loss function used here mirrors \cite{bourdinTacklingLongRangeDependencies2024}, comprising two time-domain terms, Mean Absolute Error (MAE) and Error-to-Signal Ratio (ESR), a spectral-domain term, Multi-Resolution Short Term Fourier Transform (MR-STFT) spectral loss \cite{steinmetzAuralossAudioFocused2020}, and an energy-based term, Multi-Resolution Energy Error-to-Signal Ratio (MR-EESR) \cite{bourdinTacklingLongRangeDependencies2024}.
Equations \eqref{eq:mae} and \eqref{eq:esr} define the MAE and ESR:

\vspace{-0.75em}
\begin{multicols}{2}
\noindent
\begin{equation}
\label{eq:mae}
\mathcal{L}_{\textrm{MAE}} = \frac{1}{L} \sum\limits_{t=0}^{L-1} |\hat{y}[t] - y[t]|
\end{equation}
\columnbreak
\begin{equation}
\label{eq:esr}
\mathcal{L}_{\textrm{ESR}} = \frac{{\textstyle \sum}_{t=0}^{L-1} (y[t] - \hat{y}[t])^2}{{\textstyle \sum}_{t=0}^{L-1} y[t]^2}
\end{equation}
\end{multicols}
\vspace{-2.25em}

The STFT loss, given by \eqref{eq:stft}, includes the STFT convergence and STFT magnitude terms:

\vspace{-1.5em}
\begin{equation}
    \label{eq:stft}
    \mathcal{L}_{\textrm{STFT}} = \frac{||\: |Y| - |\hat{Y}|\: ||_F}{||\:|Y|\:||_F} + \frac{1}{L} || \log(|Y|) - \log(|\hat{Y}|) ||_1
\end{equation}
\vspace{-1em}

\noindent where $Y$ and $\hat{Y}$ are the STFTs of $y$ and $\hat{y}$, and $||.||_F$ is the Frobenius norm. MR-STFT averages several STFT losses using different window sizes (512, 1024, 2048, i.e. 12, 23 and 46 ms with a sample rate of 44100 Hz).
EESR is defined in \eqref{eq:eesr}:

{ \setlength{\abovedisplayskip}{0pt} \setlength{\belowdisplayskip}{0pt}
\begin{equation}
\label{eq:eesr}
    \mathcal{L}_{\mathrm{EESR}} = \frac{1}{K} \sum\limits_{k=0}^{K-1} \frac{| \widehat{\operatorname{E}_k} - \operatorname{E}_k |}{\operatorname{E}_k} \;\; \mathrm{with} \
    \operatorname{E}_k = \frac{1}{W} \sum\limits_{\tau=kW}^{(k+1)W-1} y[\tau]^2
\end{equation}
}
\vspace{-1em}

\noindent where $W$ is the window size and $K = \lfloor L / (W/4) \rfloor$, corresponding to a 75\% overlap. MR-EESR averages several EESR losses with the same $W$ values as MR-STFT. The final loss is the sum of $100\mathcal{L}_{\textrm{MAE}}$, $\mathcal{L}_{\textrm{ESR}}$, $\mathcal{L}_{\textrm{MR-STFT}}$, and $\mathcal{L}_{\textrm{MR-EESR}}$. The factor of 100 balances the magnitude of these terms at training start.

\subsection{Dataset}
\label{dataset}

\subsubsection{Source Audio}
\label{source}

The input audio files in our study consist of several sequences. Each file starts with a 1000 Hz pure tone, where the amplitude varies in 1 dB steps from -39 dB to 0 dB. These steps are organized into four groups of ten, each lasting 0.25 seconds, and separated by -40 dB steps lasting 0.5 seconds, creating a 16-second sequence. These quieter steps trigger the release phase of the compressor.
Next, the file includes several 4-second excerpts randomly selected from the Free Music Archive dataset \cite{defferrardFMADatasetMusic2017}, which contains diverse musical tracks. Each excerpt is normalized to a maximum peak of 0 dB, and this 40-second sequence's peak amplitude decreases gradually from 0 dB to -20 dB.
The file concludes with two 20-second sequences of procedurally generated sounds. The first sequence has many sound events with few silence, while the second spaces the sound events with low-amplitude noise to capture the compressor's release phases. The sound events are created using randomized Attack-Decay-Sustain-Release envelopes modulating sounds among a white noise generator; an oscillator with three harmonics subject to random parameters and waveshaping; and finally linear and exponential chirps with randomized initial and target frequencies.

\subsubsection{Parameter Sampling}
\label{parameter}

Our work models the API-2500+ compressor, including most of its controls: Threshold, Attack, Ratio, Release, and Knee, plus the Thrust control, which enables high-pass filtering after gain detection. We focus on mono effects, so stereo controls are ignored. The compressor also has a `tone type' control, which switches between feed-forward (`new') and feed-back (`old') configurations; our study only considers feed-forward.
The parameters are discrete except the Threshold, continuous between -20 dB and +20 dB, which we discretize into 4 dB increments. We represent these controls as a vector of normalized values from 0 to 1.

Three datasets were recorded, with the described source audio but using different parameter sampling strategies. The first is a `snapshot' dataset with 16 items recorded with a single configuration of controls. Here, all controls are set to their middle values except for the release control, which is set to maximum, posing a challenging task due to the long time dependency. The second dataset is a limited full-modeling set, where only the Threshold and Ratio parameters vary while other controls remain identical to those in the snapshot dataset. This dataset also contains 16 items, with Threshold values at 4, 0, -4, and -8 dB, and Ratio values at 3, 4, 6, and 10. A third dataset is the full-modeling dataset considered in \cite{bourdinTacklingLongRangeDependencies2024}, containing 160 items. Parameters in this dataset were sampled using Latin Hypercube Sampling, to facilitate a unique train-validation-test split, ensuring that each subset was representative of the control parameter distributions.

The recordings were made using an Arturia AudioFuse sound card, at a sample rate of 44100 Hz. The output of the sound card, which is the compressor input, is looped back to the sound card. The slight DC biases of the card's inputs are removed in post-processing by substracting the mean out of each recording.

\subsubsection{Cross-Validation}
\label{cross}

In supervised learning, the goal is to train a model that performs accurately on unseen data from the same distribution as the training dataset, minimizing the expected risk \(\mathbb{E}_{(x,y)\sim \mathcal{D}}[e(h(x), y)]\), where \(\mathcal{D}\) is the data distribution, \(e\) is a metric, and \(h\) is the model \cite{bouthillierAccountingVarianceMachine2021}. The standard practice is dividing the dataset into three subsets: training, used by the optimization algorithm; validation, to monitor generalization and prevent overfitting; and test, for final evaluation.
Selecting these subsets is essentially a realization of a random variable, as it involves sampling from the full data distribution \(\mathcal{D}\). The training pipeline also faces variance from sources like initial model weights and batches constitution and order. Thus, evaluating a single model instance should be seen as a realization of a random variable \cite{bouthillierAccountingVarianceMachine2021}, and training a model once does not reliably estimate risk or allow meaningful model comparisons. Cross-validation offers a better risk estimate by training multiple instances on different splits. However, due to the high training cost of deep networks, most studies train models once on a single split.
In this work, we extracted 10 train-validation-test splits from each dataset by shuffling items before splitting. For the snapshot and Threshold-Ratio datasets, each with 16 items, we shuffled and split the items into 8 training, 4 validation, and 4 test items, repeating 10 times. For the third dataset (full-modeling) with 160 items, the list was split into 128 training, 16 validation, and 16 test items.

\subsection{Training}
\label{training}

\subsubsection{Temporal Operations and Cached Padding}
\label{temporal}

In convolutional neural networks, operations like convolutions and pooling alter the temporal length of inputs. Convolutions with kernel size $k$ (dilation and stride 1) reduce input length by $k-1$ samples, while pooling divides it by a factor.
Though shorter outputs help in tasks like classification, audio effect modeling needs the model output to match the input length. Zero-padding is a common solution to this mismatch but can complicate inference, causing discontinuities at the start of output buffers when processing sequential sample buffers. To solve this, \cite{caillonStreamableNeuralAudio2022Dafx} suggests cached padding, storing the last samples of an input tensor in cache before applying a temporal operation, so these samples can be used for padding in subsequent operations. 

Cached padding is effective with consecutive buffers but still relies on zero-padding when the cache is not initialized, e.g. at the start of a training batch. Using zeros instead of actual sample values can introduce bias, especially when large numbers of zeros are used relative to the input length. Moreover, models like Temporal Convolutional Networks (TCNs) \cite{steinmetzEfficientNeuralNetworks2022} with large receptive fields may compute outputs predominantly from zeros rather than true signal values, necessitating large sequence lengths in training batches.  In \cite{bourdinTacklingLongRangeDependencies2024}, zero-padding was avoided by calculating input samples required by a model to achieve the desired output length.

Models with binary operations involving 1D tensors, like skip connections, need identical tensor lengths. In SPTMod, this occurs with TFiLM operations following each ModBlock. Without padding, tensors often differ in length, necessitating cropping layers to trim the longer tensor (on its left-hand side, for causality) to match the shorter one. Determining input lengths and cropping sizes presents an optimization challenge. The goal is to minimize cropping sizes (which must be non-negative) while ensuring that the input tensor lengths of the binary operations are equal, and that the input lengths of pooling layers are multiples of the pooling size. An optimization algorithm can solve this for any architecture and fixed hyperparameters. For architectures like SPTMod, we derived an optimal solution, detailed at our accompanying repository, depending on the pooling size and the convolution hyperparameters.

\subsubsection{Windowed Target and Streamed Target}
\label{windowed}

The use of models with recurrent layers varies between training and inference. During training, models often process nonconsecutive batches of sequences, as datasets typically consist of long sequences that are sliced and shuffled during training. Consequently, internal states may not persist across iterations. The Windowed Target (WT) error \cite{hawleySignalTrainProfilingAudio2019} measures a metric (e.g., loss) like during training. In contrast, during inference, models still process inputs in a windowed fashion, yet states are not reset between batches, corresponding to the Streamed Target (ST) setting. The ST error provides a more accurate user-observed metric. Ideally, minimizing WT error during training should generalize to minimizing ST error during inference, especially via techniques like warm-up or SPNs.
However, previous work \cite{bourdinTacklingLongRangeDependencies2024} showed ST error can increase compared to WT error when using state prediction. It was hypothesized this is due to exposure bias: models do not learn to recover from erroneous state values. Two strategies are possible to reduce reliance on the SPN, mitigating exposure bias: resetting states less often, and/or avoiding SPN overfitting. The first strategy can be done by increasing sequence length, though this raises computational costs and reduces weight update frequency. The second strategy uses regularization techniques like dropout to prevent overfitting state predictions.

\subsubsection{Truncated Backpropagation Through Time}
\label{truncated}

\begin{figure}[t]
\begin{center}
\includegraphics[width=0.95\columnwidth]{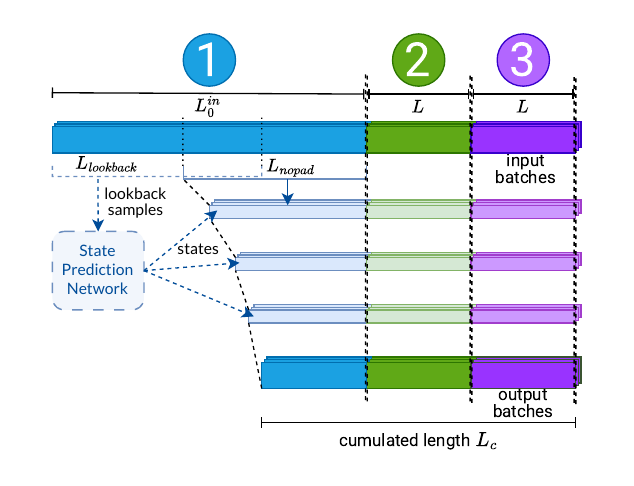}
\caption{
Diagram of intermediary tensor lengths for consecutive (non-overlapping) sequence batches in our TBPTT-based approach with $N=3$. In the first iteration, no padding is applied, so the input length includes the samples needed for temporal operations. In subsequent iterations, states and caches are retained, but their gradients are detached from the computational graph.
}
\vspace{-2em}
\label{fig:optstrat}
\end{center}
\end{figure}

Recurrent networks can be trained by updating weights after each processed time step, but this is computationally expensive and limits the loss function to sample-to-sample differences. More commonly, recurrent networks process full sequences before updating weights with Backpropagation Through Time (BPTT) \cite{werbos1990backpropagation}. However, very long sequences result in infrequent weight updates. 

Truncated BPTT (TBPTT) \cite{elman1990finding} improves efficiency by splitting long sequences into smaller ones, updating weights after processing each smaller sequence while preserving recurrent layer states.

We propose an optimization strategy based on TBPTT, detailed in Figure \ref{fig:optstrat}. This strategy introduces two hyperparameters: \(N\), number of sub-sequences, and \(L\), sub-sequence length, allowing us to define the cumulative length \(L_c\) as \(L_c = N \cdot L\). The strategy divides training into groups of \(N\) iterations, processing a cumulative sequence with TBPTT. In the first iteration, padding is not used, so the input length \(L_{\text{nopad}}\) must be adjusted to achieve an output length equal to \(L\). The SPN initializes states, requiring \(L_{\text{lookback}}\) samples. Thus, the input buffer length for the first iteration is \(L_{\text{in}}^0 = L + \max(L_{\text{nopad}} - L, L_{\text{lookback}})\). The SPN receives the first \(L_{\text{lookback}}\) samples, while the effect processor network receives the last \(L_{\text{nopad}}\) samples and generates \(L\) output samples. In subsequent TBPTT iterations, the SPN is not used and cached padding is employed, making input length equal to output length \(L\).

To form batches, the dataset is divided into long sequences of length \(L_{\text{in}}^0 + (N - 1) \cdot L\) with a step of \(L_c\). At the start of each epoch, sequences are shuffled and batches formed. These batches, containing long sequences, are further sliced into \(N\) batches of shorter consecutive sequences of lengths \((L_{\text{in}}^0, L, ..., L)\). Every \(N\) iterations, states and caches are reset, since a new batch from a different long sequence is processed.

\subsubsection{Training Procedure}
\label{trainingprocedure}

The models and their training are implemented using the PyTorch deep learning library. We use the Adam optimizer with its default parameters and a learning rate set to $5\times 10^{-4}$. Due to varying batch sizes and sequence lengths in this work, the number of batches per epoch also varies. At the end of each epoch, we measure the validation loss in both WT and ST modes. %However, the primary focus is on monitoring the ST validation loss. 
We use early stopping, halting training if the ST validation loss does not improve over $76800$ iterations, with a cap of 1 million iterations.

\subsubsection{Hyperparameters}
\label{hyperparameters}

SPTMod involves hyperparameters which are the number of modulation blocks and, within each block, parameters related to the convolutional layer: the number of channels, kernel size, and dilation size. Additional hyperparameters include the pooling size, LSTM hidden size, and the number of hidden neurons in the FiLM conditioning layer. 

In \cite{bourdinTacklingLongRangeDependencies2024}, a hyperparameter search led to a setup with 3 modulation blocks. The convolutional layer in each block has 21, 19, and 32 output channels respectively, with kernel sizes of 9, 29, and 25. The pooling size is set to 95, the LSTM hidden size is 31, and the FiLM hidden layers have 26 neurons. The SPN has 7 blocks with 16 channels, a kernel size of 38, a pooling size of 4, and a hidden layer size of 8 in FiLM, achieving a lookback length of $\sim$5 seconds. This model configuration is called SPTMod24. Note that within our framework, the pooling size must be a divisor of the sequence length; thus, we set the pooling size to 64 instead of 95 in SPTMod24. 

In this study, we explore different hyperparameters in a model called SPTMod25. This model has 4 modulation blocks, each with 15 output channels and a kernel size of 3 in the convolutional layers, and 32 neurons in the FiLM hidden layers for both modulation and SPN blocks. The other hyperparameters are identical to those used in SPTMod24. This configuration was chosen to investigate the impact of model depth over width, hypothesizing that increasing the number of blocks to 4 with simpler hyperparameter choices could offer similar or improved performance without the computational cost of an extensive hyperparameter search. The shrinkage in model width (number of channels) led to roughly a 10x reduction in multiply/add operations.

\section{Experiments and Results}
\label{experiments}
\subsection{Adjusting Truncated Backpropagation Through Time}
\label{adjusting}

TBPTT highlights two key hyperparameters: number of sequences \(N\) and sequence length \(L\). The choice between BPTT or TBPTT, and specific values for \(N\) and \(L\), is empirical and varies across studies modeling audio effects with recurrent layers. For instance, \cite{wright2019real} and \cite{wrightGreyBoxModellingDynamic2022} use TBPTT with warm-up lengths of 1000 and 4096 samples, and sequence lengths of 2048 and 8128 samples, achieving cumulative lengths of 0.5s and 2.5s, $N=10$ and 13. Similarly, \cite{mikkonenSamplingUserControls2024} uses TBPTT with both warm-up and sequence lengths of 1024 samples, resulting in a cumulative length of 1s and \(N=42\). In contrast, \cite{ramirezDeepLearningAudio2021} and \cite{comunitaModellingBlackBoxAudio2023} do not use TBPTT or warm-up, with sequence lengths from 1024 to 8192 samples in \cite{ramirezDeepLearningAudio2021} and set to 112640 in \cite{comunitaModellingBlackBoxAudio2023}.

In our preliminary testing, we assessed a grid of \(N\) and \(L\) values, maintaining a fixed batch size of 16 and using the snapshot dataset. We found that training a neural effect in snapshot modeling led to faster and more stable convergence, with less variance than in full modeling. For smaller \(L\) values (2048-8192), the loss was notably high, likely due to low information content in a batch. Increasing the batch size by a factor of four resulted in a final loss closer to what was observed with larger \(L\) values. Based on this, we included batch size \(B\) as a hyperparameter in further experiments.
The TBPTT experiment presented here includes both the snapshot and Threshold-Ratio datasets, to determine if the effects of these hyperparameters differ between snapshot and full modeling. We considered the Cartesian product of \(N \in \{1, 2, 3\}\), \(B \in \{8, 16, 32, 64, 128\}\), and \(L \in \{4096, 8192, 16384, 32768\}\), excluding \((B, L)=(128, 32768)\) due to excessive memory requirements that could not be met by all nodes of our heterogeneous computing platform. For each dataset and each combination of \(N\), \(B\), and \(L\), we trained 10 model instances, one for each train-validation-test split described in Section \ref{cross}.
Due to a substantial 570 training runs per dataset, we limited the experiment to the first two datasets, excluding the large full-modeling one.

\subsubsection*{Results}

\renewcommand\tabularxcolumn[1]{m{#1}}
\renewcommand{\arraystretch}{0.75}
\input{table_median_snap}
\input{table_median_threshratio}
\input{table_mad_snap}
\input{table_mad_threshratio}
\input{table_time_median_snap}
\input{table_time_median_threshratio}
% \afterpage{\afterpage{\cleardoublepage}}

The main results of this experiment are shown in tables (1-6), which detail statistics on validation losses and training times for both the snapshot and Threshold-Ratio datasets across all considered hyperparameter configurations.
Tables \ref{tab:loss_median_snap} and \ref{tab:loss_median_threshratio} present median validation loss values to estimate expected model accuracy under given $(N, B, L)$ hyperparameters. Generally, median validation loss improves as $N$, $B$, or $L$ increase. On both datasets, increasing batch size has a strong impact at low $L_c$ values, but this effect fades as $L_c$ rises, especially on the snapshot dataset when $L_c \geq 32768$. On the Threshold-Ratio dataset, batch size has more impact. Increasing $L$ consistently enhances accuracy, while $N$ tends to do so. Sometimes, $N=1$ (i.e., without TBPTT) results in a lower median validation loss than higher $N$ values.

\noindent Tables \ref{tab:loss_mad_snap} and \ref{tab:loss_mad_threshratio} show the Median Absolute Deviation (MAD) values (defined as the median of the absolute deviations from the data's median) of validation losses to estimate model training stability under specific $(N, B, L)$. Although trends are harder to discern due to limited values for proper MAD evaluation, using TBPTT ($N > 1)$ typically improves MAD values. However, for high $L$ (thus so is $L_c$), small MAD values are achieved with $N=1$.

Tables \ref{tab:time_median_snap} and \ref{tab:time_median_threshratio} provide median training times for each configuration. Training duration is calculated by determining the number of iterations needed to reach the minimum validation loss within a 5\% margin, then multiplying by the seconds per iteration factor. This factor was pre-evaluated in all configurations on a unique computer with an NVIDIA L40 GPU. Higher $B$ and $L$ values generally increase training time due to more computation per batch, but convergence can occur in fewer iterations and less time, as shown when $B=8$ and $L$ rises in Table \ref{tab:time_median_threshratio}. Increasing $N$ often significantly improves training times because the SPN is invoked only at one iteration out of $N$.

\subsubsection*{Discussion}

The choice of $(N, B, L)$ greatly impacts the training pipeline performance, affecting accuracy, time, and stability. As predicted in \cite{mccandlish2018empirical}, larger batch sizes surpassing a critical value yield diminishing returns, and the more complex the modeling task, the higher the critical batch size. For the snapshot dataset, this critical size is low, while it's high in full modeling -- more than expected -- as shown by the Threshold-Ratio dataset with 2 control parameters. Thus, one might expect even higher ideal batch sizes for full modeling datasets with more control parameters, like ours with 7 controls. A relevant compromise may be using large batch sizes with low sequence lengths and high $N$ values. Training variance, measured by MAD values, is generally greater on the Threshold-Ratio dataset than the snapshot one, likely due to increased data diversity with more control parameters. Lastly, and unexpectedly, training time is not significantly shorter with snapshot than full modeling.

\subsection{Objective Evaluation}
\label{objective}

Following the TBPTT experiment, SPTMod24 and SPTMod25 were trained and assessed on the large full-modeling dataset, under 3 specific hyperparameter sets for comparison: $(N, B, L)$ = (1, 16, 16384), mirroring \cite{bourdinTacklingLongRangeDependencies2024}, and two top-performing configurations, $(N, B, L)$ = (1, 64, 32768) and $(N, B, L)$ = (3, 64, 32768).

%\subsubsection*{Results}

\begin{figure}[htb]
\begin{center}
\includegraphics[width=0.95\columnwidth]{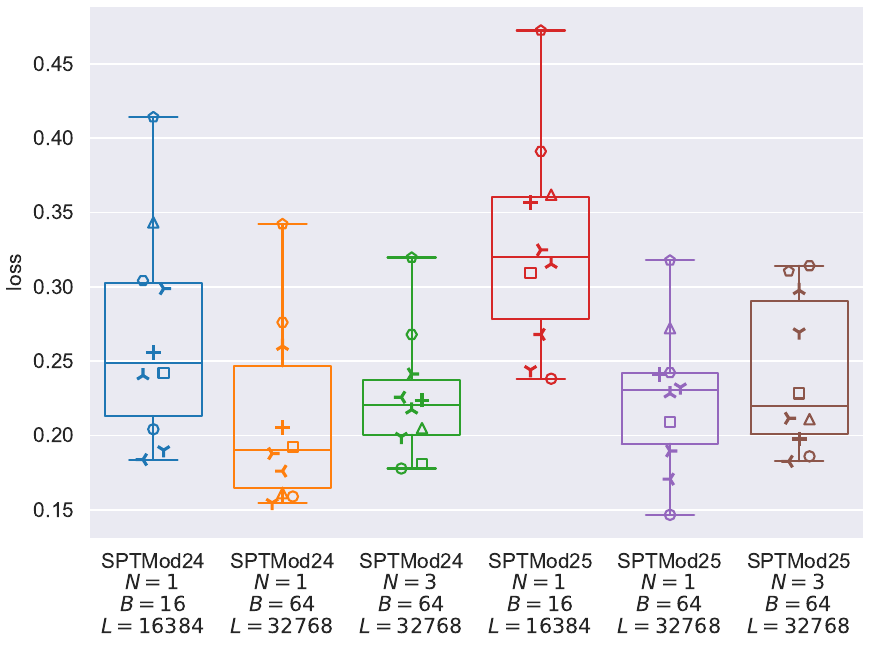}
\caption{Loss after training the 6 models on 10 splits each, evaluated on their respective test subsets, depicted by markers.}
\label{fig:objeval}

\end{center}
\end{figure}

Figure \ref{fig:objeval} shows all test loss values for the models considered. Under the same $(N,B,L)$ values as in \cite{bourdinTacklingLongRangeDependencies2024}, where $(N,B,L)$ = (1, 16, 16384), SPTMod25 performs worse than SPTMod24. However, with new training hyperparameters, SPTMod24 and SPTMod25 perform similarly, though SPTMod24 with $(N,B,L)$ = (1, 64, 32768) appears to outperform the other models. 
We conclude from this objective evaluation that training hyperparameters $N$, $B$ and $L$ are at least as crucial as architectural ones.

\subsection{Subjective Evaluation}
\label{subjective}

% \begingroup
\setlength{\intextsep}{0em}%
\setlength{\columnsep}{0.8em}%

\begin{wrapfigure}{r}{0.37\linewidth}
    \centering
    \includegraphics[width=0.82\linewidth]{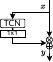}
    \caption{TCN variant in the listening test}
    \label{fig:tcn_gr}
\end{wrapfigure}

A listening test was designed to assess the perceptual quality of the models. 
For the subjective evaluation, we selected (1) \mbox{SPTMod24} with \((N, B, L)=(1, 16, 16384)\), as featured in \cite{bourdinTacklingLongRangeDependencies2024}; (2) SPTMod25 with \((N,B,L) = (1, 64, 32768)\); and (3) a variant of a Temporal Convolutional Network (TCN) model \cite{steinmetzEfficientNeuralNetworks2022}, shown in Figure \ref{fig:tcn_gr}, with \((N,B,L) = (1, 64, 32768)\). 
This TCN variation computes a gain reduction signal applied to the input. It serves as a non-recurrent baseline, thus with no WT/ST difference, and is more comparable to our model.

We followed the MUSHRA method (MUlti Stimulus test with Hidden Reference and Anchor) \cite{series2014method}, a listening test where participants rate the perceptual quality of multiple excerpts on a scale from 0 to 100, relative to a reference: the target effect's output. The test was conducted online via a webMUSHRA instance \cite{schoeffler2018webmushra}.\footnote{Exposed at \url{https://ybourdin.github.io/sptmod/}} 
It includes outputs from the three models, a hidden reference identical to the provided reference, and an anchor processed through a rough compressor implementation. 
The hidden reference and anchor assess participant reliability and their ability to distinguish excerpts.
The test featured 3 input audio excerpts: drumming, guitar strumming, and gypsy guitar playing. We evaluated 4 compressor configurations: low attack, low attack and release, high release, and high `thrust'. Each audio and control combination was evaluated twice: once on a 4-beat excerpt (about 4 seconds) and once on a 1-beat segment (about 1 second) isolated from the excerpt. This approach ensures participants make decisions based on the same content.
In total, 24 evaluation items were assessed. To reduce test duration, items were split into two groups of 12, with each participant receiving a random group and order, except that a 4-beat excerpt and its 1-beat version remain consecutive. 
Though the test was designed to last 15 minutes, participants completed the test with unlimited time and repetitions, and the median total rating time is 26 minutes.
Afterward, they reported their audio device and answered questions about their expertise, particularly whether they had used a dynamic range compressor or had mixing or mastering skills. Of the 29 participants, 14 met the selection criteria for final results: positive responses to at least one expertise question, use of headphones or monitoring speakers, no more than three ratings below 90 for the hidden reference, and no more than two ratings above 90 for the anchor. This yielded 168 item evaluations.

% \endgroup

\subsubsection*{Results}

\begin{figure}[htb]
\begin{center}
\captionsetup{belowskip=0pt}  % Diminue l'espace sous la légende
\includegraphics[width=0.88\columnwidth]{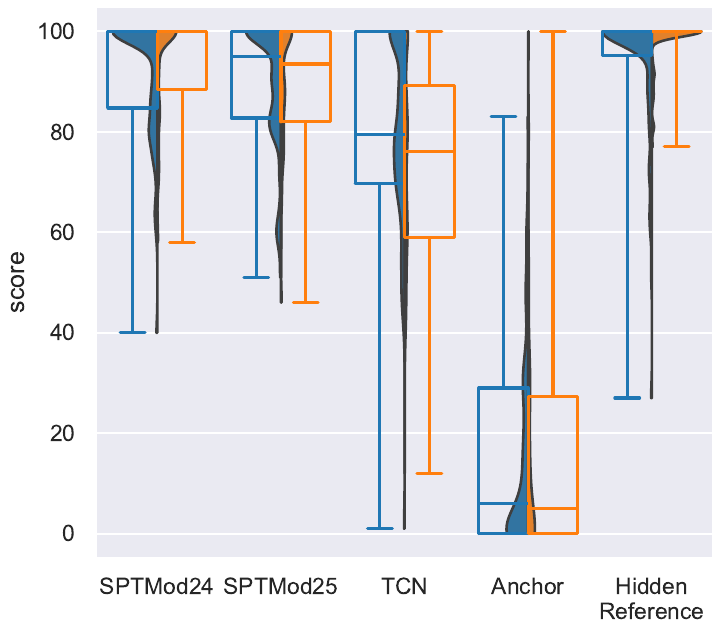}
\caption{Scores of the listening test, split into the blue (longer loops --- 4 beats) and the orange (shorter loops -- 1 beat) groups.}
\label{fig:mushra_violinall}
\end{center}
\end{figure}

\begin{figure*}[htb]
    \captionsetup[subfigure]{aboveskip=-5pt,belowskip=-5pt}
    \centering
    \begin{subfigure}[b]{0.38\textwidth}
        \centering
        \includegraphics[width=\textwidth]{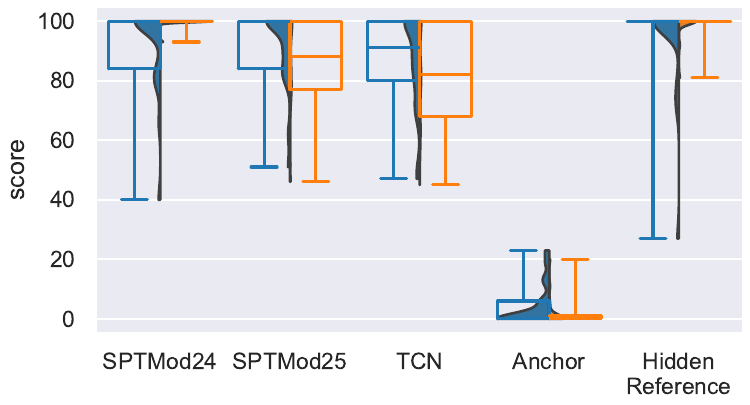}
        \caption[]%
        {{\small High attack time.}}    
        \label{fig:mushra_0}
    \end{subfigure}
    \hspace{68pt} % au doigt mouillé pour centrer par rapport aux colonnes
    \begin{subfigure}[b]{0.38\textwidth}  
        \centering 
        \includegraphics[width=\textwidth]{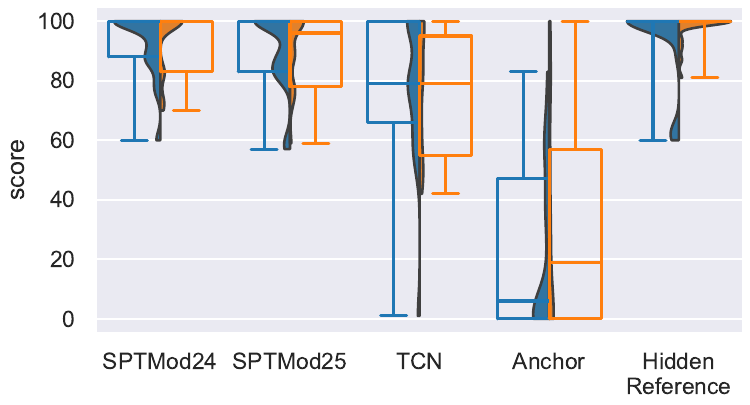}
        \caption[]%
        {{\small Low attack and release times.}}    
        \label{fig:mushra_1}
    \end{subfigure}
    \vskip\baselineskip
    \begin{subfigure}[b]{0.38\textwidth}   
        \centering 
        \includegraphics[width=\textwidth]{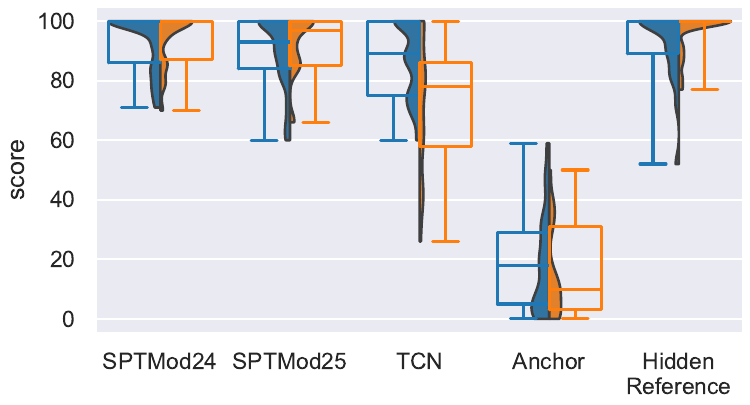}
        \caption[]%
        {{\small High release time.}}    
        \label{fig:mushra_2}
    \end{subfigure}
    \hspace{68pt}
    \begin{subfigure}[b]{0.38\textwidth}   
        \centering 
        \includegraphics[width=\textwidth]{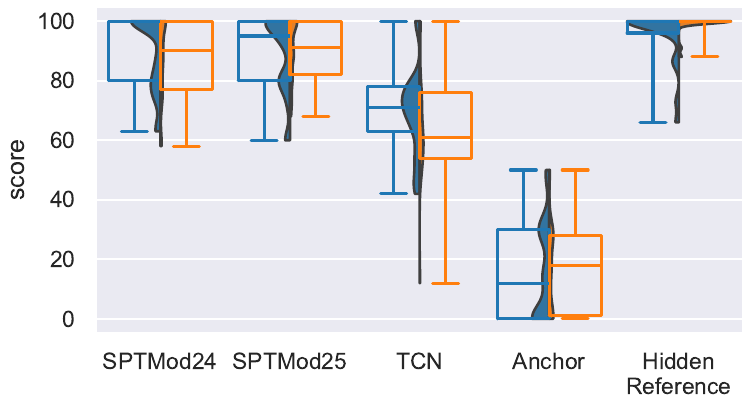}
        \caption[]%
        {{\small High Thrust.}}    
        \label{fig:mushra_3}
    \end{subfigure}
    \vspace{5pt}
    \caption[]
    {Scores of the listening test, grouped by parameter configuration, split into the blue (longer loops -- 4 beats) and the orange (shorter loops -- 1 beat) groups.} 
    \label{fig:mushra}
\end{figure*}

Figure \ref{fig:mushra_violinall} shows statistics on scores given to each model. The median score of SPTMod24 is 100, indicating this model, trained with the hyperparameters in \cite{bourdinTacklingLongRangeDependencies2024}, achieves excellent perceptual quality. SPTMod25 is slightly worse, with a median score between 90 and 95, which is satisfying given its smaller size and straightforward tuning. The TCN variant is the worst, still providing good quality but being easier to discern.
As expected, the scores assigned to the hidden references among shorter loops (in orange) have lower variability than longer ones. Thus, more confidence should be put in the scores given on shorter loops.
In Figure \ref{fig:mushra}, scores are grouped by the parameter configurations described earlier, aligning with the overall mean results. On the high attack time configuration (\ref{fig:mushra_0}), SPTMod24 achieves outstanding scores, while in the other configurations (\ref{fig:mushra_1}, \ref{fig:mushra_2}, \ref{fig:mushra_3}) they are very high and similar to SPTMod25. Although TCN performs worse overall, it still achieves relatively high scores across most configurations, particularly in the high release setting, though it struggles with the high thrust configuration.

\section{Conclusion}
\label{conclusion}

This study investigated the parameters of TBPTT in the context of training a convolutional-recurrent network for modeling digital audio effects, focusing on DRC. Our main contribution is the empirical evaluation of TBPTT hyperparameters -- namely, the number of sequences, batch size, and sequence length -- and their effects on model accuracy, training stability, and computational efficiency.
Extensive tests with both snapshot and full-modeling datasets show that increasing these hyperparameters generally improves model accuracy and reduces training variability, though it raises computational demands. 
Notably, TBPTT lowers validation loss variance and speeds up training, potentially allowing for larger batch sizes, which is beneficial for modeling multiple control parameters. These findings suggest that those hyperparameters are as crucial as architectural choices for optimal performance.
Objective evaluation confirms that models trained with newly optimized TBPTT parameters perform better than prior setups. However, a subjective listening test revealed that even previously, SPTMod24 had excellent perceptual quality, indicating the revised TBPTT setup did not significantly impact the model's perceptual quality.
For future work, implementing the model in C++ will allow for a precise estimation of its computational efficiency and real-time viability, as well as enabling multi-objective hyperparameter search that considers both model accuracy and computational cost.

\section{Acknowledgments}

This work is part of a Cifre PhD project funded by ANRT. \sloppy{It benefited from access to the computing resources of the `PLaFRIM' and `CALI 3' clusters. PLaFRIM is supported by Inria, CNRS (LABRI \& IMB), Université de Bordeaux, Bordeaux INP and Conseil Régional d'Aquitaine. CALI 3 is operated by the University of Limoges and is part of the HPC network in the Nouvelle-Aquitaine Region, financed by the State and the Region.}

%\newpage
% \nocite{*} % ce truc fait afficher TOUTE la biblio du fichier
\bibliographystyle{IEEEbib}
% \bibliography{bourdin} % requires file DAFx25_tmpl.bib

\input{bourdin.bbl}
\end{document}

%% file: table_median_snap.tex
\begin{table*}[p]
\centering
\begin{subtable}[t]{0.29\textwidth}
\centering
\small \begin{tabularx}{\linewidth}{{lXXXX}}
\toprule
\makebox[25pt][r]{$L$} & \makebox[0.5\linewidth][c]{4096} & \makebox[0.8\linewidth][c]{8192} & \makebox[1.1\linewidth][c]{16384} & \makebox[1.4\linewidth][c]{32768} \\
$B$ \textbackslash~ {\color{gray} $L_c$} & \makebox[0.5\linewidth][c]{\color{gray}4096} & \makebox[0.8\linewidth][c]{\color{gray}8192} & \makebox[1.1\linewidth][c]{\color{gray}16384} & \makebox[1.4\linewidth][c]{\color{gray}32768} \\
\midrule
8 & \makebox[\linewidth]{{\cellcolor[HTML]{EB5A3A}} \color[HTML]{F1F1F1} \raisebox{-1pt}{\small 3.70}} & \makebox[\linewidth]{{\cellcolor[HTML]{FA9B58}} \color[HTML]{000000} \raisebox{-1pt}{\small 3.18}} & \makebox[\linewidth]{{\cellcolor[HTML]{FAFDC9}} \color[HTML]{000000} \raisebox{-1pt}{\small 2.20}} & \makebox[\linewidth]{{\cellcolor[HTML]{78B0D3}} \color[HTML]{F1F1F1} \raisebox{-1pt}{\small 1.46}} \\
16 & \makebox[\linewidth]{{\cellcolor[HTML]{F57245}} \color[HTML]{F1F1F1} \raisebox{-1pt}{\small 3.48}} & \makebox[\linewidth]{{\cellcolor[HTML]{FFF6B1}} \color[HTML]{000000} \raisebox{-1pt}{\small 2.36}} & \makebox[\linewidth]{{\cellcolor[HTML]{9DCEE3}} \color[HTML]{000000} \raisebox{-1pt}{\small 1.61}} & \makebox[\linewidth]{{\cellcolor[HTML]{35429B}} \color[HTML]{F1F1F1} \raisebox{-1pt}{\small 1.10}} \\
32 & \makebox[\linewidth]{{\cellcolor[HTML]{A50026}} \color[HTML]{F1F1F1} \raisebox{-1pt}{\small 4.77}} & \makebox[\linewidth]{{\cellcolor[HTML]{FDBB6D}} \color[HTML]{000000} \raisebox{-1pt}{\small 2.92}} & \makebox[\linewidth]{{\cellcolor[HTML]{588CC0}} \color[HTML]{F1F1F1} \raisebox{-1pt}{\small 1.32}} & \makebox[\linewidth]{{\cellcolor[HTML]{3E60AA}} \color[HTML]{F1F1F1} \raisebox{-1pt}{\small 1.18}} \\
64 & \makebox[\linewidth]{{\cellcolor[HTML]{FCA85E}} \color[HTML]{000000} \raisebox{-1pt}{\small 3.09}} & \makebox[\linewidth]{{\cellcolor[HTML]{FEEBA1}} \color[HTML]{000000} \raisebox{-1pt}{\small 2.49}} & \makebox[\linewidth]{{\cellcolor[HTML]{5C90C2}} \color[HTML]{F1F1F1} \raisebox{-1pt}{\small 1.33}} & \makebox[\linewidth]{{\cellcolor[HTML]{416AAF}} \color[HTML]{F1F1F1} \raisebox{-1pt}{\small 1.21}} \\
128 & \makebox[\linewidth]{{\cellcolor[HTML]{FEE294}} \color[HTML]{000000} \raisebox{-1pt}{\small 2.58}} & \makebox[\linewidth]{{\cellcolor[HTML]{FFF8B5}} \color[HTML]{000000} \raisebox{-1pt}{\small 2.32}} & \makebox[\linewidth]{{\cellcolor[HTML]{8EC2DC}} \color[HTML]{000000} \raisebox{-1pt}{\small 1.55}} &  \\
\bottomrule
\end{tabularx}
\captionsetup{skip=2pt}
\caption{N = 1}
\end{subtable}
\hfill
\begin{subtable}[t]{0.29\textwidth}
\centering
\small \begin{tabularx}{\linewidth}{{lXXXX}}
\toprule
\makebox[25pt][r]{$L$} & \makebox[0.5\linewidth][c]{4096} & \makebox[0.8\linewidth][c]{8192} & \makebox[1.1\linewidth][c]{16384} & \makebox[1.4\linewidth][c]{32768} \\
$B$ \textbackslash~ {\color{gray} $L_c$} & \makebox[0.5\linewidth][c]{\color{gray}8192} & \makebox[0.8\linewidth][c]{\color{gray}16384} & \makebox[1.1\linewidth][c]{\color{gray}32768} & \makebox[1.4\linewidth][c]{\color{gray}65536} \\
\midrule
8 & \makebox[\linewidth]{{\cellcolor[HTML]{A90426}} \color[HTML]{F1F1F1} \raisebox{-1pt}{\small 4.70}} & \makebox[\linewidth]{{\cellcolor[HTML]{FEE99D}} \color[HTML]{000000} \raisebox{-1pt}{\small 2.52}} & \makebox[\linewidth]{{\cellcolor[HTML]{81B7D7}} \color[HTML]{000000} \raisebox{-1pt}{\small 1.49}} & \makebox[\linewidth]{{\cellcolor[HTML]{4471B2}} \color[HTML]{F1F1F1} \raisebox{-1pt}{\small 1.23}} \\
16 & \makebox[\linewidth]{{\cellcolor[HTML]{FDAF62}} \color[HTML]{000000} \raisebox{-1pt}{\small 3.03}} & \makebox[\linewidth]{{\cellcolor[HTML]{FEEFA6}} \color[HTML]{000000} \raisebox{-1pt}{\small 2.44}} & \makebox[\linewidth]{{\cellcolor[HTML]{70A9CF}} \color[HTML]{F1F1F1} \raisebox{-1pt}{\small 1.42}} & \makebox[\linewidth]{{\cellcolor[HTML]{4D7FB9}} \color[HTML]{F1F1F1} \raisebox{-1pt}{\small 1.27}} \\
32 & \makebox[\linewidth]{{\cellcolor[HTML]{FEE99D}} \color[HTML]{000000} \raisebox{-1pt}{\small 2.51}} & \makebox[\linewidth]{{\cellcolor[HTML]{F0F9DB}} \color[HTML]{000000} \raisebox{-1pt}{\small 2.10}} & \makebox[\linewidth]{{\cellcolor[HTML]{78B0D3}} \color[HTML]{F1F1F1} \raisebox{-1pt}{\small 1.46}} & \makebox[\linewidth]{{\cellcolor[HTML]{313695}} \color[HTML]{F1F1F1} \raisebox{-1pt}{\small 1.07}} \\
64 & \makebox[\linewidth]{{\cellcolor[HTML]{FECE7F}} \color[HTML]{000000} \raisebox{-1pt}{\small 2.76}} & \makebox[\linewidth]{{\cellcolor[HTML]{D1ECF4}} \color[HTML]{000000} \raisebox{-1pt}{\small 1.86}} & \makebox[\linewidth]{{\cellcolor[HTML]{4878B6}} \color[HTML]{F1F1F1} \raisebox{-1pt}{\small 1.25}} & \makebox[\linewidth]{{\cellcolor[HTML]{333D99}} \color[HTML]{F1F1F1} \raisebox{-1pt}{\small 1.09}} \\
128 & \makebox[\linewidth]{{\cellcolor[HTML]{FEDA8A}} \color[HTML]{000000} \raisebox{-1pt}{\small 2.67}} & \makebox[\linewidth]{{\cellcolor[HTML]{A8D6E8}} \color[HTML]{000000} \raisebox{-1pt}{\small 1.65}} & \makebox[\linewidth]{{\cellcolor[HTML]{5E93C3}} \color[HTML]{F1F1F1} \raisebox{-1pt}{\small 1.34}} &  \\
\bottomrule
\end{tabularx}
\captionsetup{skip=2pt}
\caption{N = 2}
\end{subtable}
\hfill
\begin{subtable}[t]{0.29\textwidth}
\centering
\small \begin{tabularx}{\linewidth}{{lXXXX}}
\toprule
\makebox[25pt][r]{$L$} & \makebox[0.5\linewidth][c]{4096} & \makebox[0.8\linewidth][c]{8192} & \makebox[1.1\linewidth][c]{16384} & \makebox[1.4\linewidth][c]{32768} \\
$B$ \textbackslash~ {\color{gray} $L_c$} & \makebox[0.5\linewidth][c]{\color{gray}12288} & \makebox[0.8\linewidth][c]{\color{gray}24576} & \makebox[1.1\linewidth][c]{\color{gray}49152} & \makebox[1.4\linewidth][c]{\color{gray}98304} \\
\midrule
8 & \makebox[\linewidth]{{\cellcolor[HTML]{F46D43}} \color[HTML]{F1F1F1} \raisebox{-1pt}{\small 3.54}} & \makebox[\linewidth]{{\cellcolor[HTML]{FEDA8A}} \color[HTML]{000000} \raisebox{-1pt}{\small 2.67}} & \makebox[\linewidth]{{\cellcolor[HTML]{A3D3E6}} \color[HTML]{000000} \raisebox{-1pt}{\small 1.64}} & \makebox[\linewidth]{{\cellcolor[HTML]{4D7FB9}} \color[HTML]{F1F1F1} \raisebox{-1pt}{\small 1.28}} \\
16 & \makebox[\linewidth]{{\cellcolor[HTML]{FED889}} \color[HTML]{000000} \raisebox{-1pt}{\small 2.68}} & \makebox[\linewidth]{{\cellcolor[HTML]{DAF0F6}} \color[HTML]{000000} \raisebox{-1pt}{\small 1.90}} & \makebox[\linewidth]{{\cellcolor[HTML]{5183BB}} \color[HTML]{F1F1F1} \raisebox{-1pt}{\small 1.29}} & \makebox[\linewidth]{{\cellcolor[HTML]{3A51A2}} \color[HTML]{F1F1F1} \raisebox{-1pt}{\small 1.14}} \\
32 & \makebox[\linewidth]{{\cellcolor[HTML]{FEDA8A}} \color[HTML]{000000} \raisebox{-1pt}{\small 2.67}} & \makebox[\linewidth]{{\cellcolor[HTML]{CFEBF3}} \color[HTML]{000000} \raisebox{-1pt}{\small 1.85}} & \makebox[\linewidth]{{\cellcolor[HTML]{4B7DB8}} \color[HTML]{F1F1F1} \raisebox{-1pt}{\small 1.27}} & \makebox[\linewidth]{{\cellcolor[HTML]{4D7FB9}} \color[HTML]{F1F1F1} \raisebox{-1pt}{\small 1.27}} \\
64 & \makebox[\linewidth]{{\cellcolor[HTML]{FDC778}} \color[HTML]{000000} \raisebox{-1pt}{\small 2.83}} & \makebox[\linewidth]{{\cellcolor[HTML]{C7E7F1}} \color[HTML]{000000} \raisebox{-1pt}{\small 1.81}} & \makebox[\linewidth]{{\cellcolor[HTML]{4878B6}} \color[HTML]{F1F1F1} \raisebox{-1pt}{\small 1.25}} & \makebox[\linewidth]{{\cellcolor[HTML]{384CA0}} \color[HTML]{F1F1F1} \raisebox{-1pt}{\small 1.13}} \\
128 & \makebox[\linewidth]{{\cellcolor[HTML]{FEECA2}} \color[HTML]{000000} \raisebox{-1pt}{\small 2.47}} & \makebox[\linewidth]{{\cellcolor[HTML]{E2F4F4}} \color[HTML]{000000} \raisebox{-1pt}{\small 1.96}} & \makebox[\linewidth]{{\cellcolor[HTML]{4878B6}} \color[HTML]{F1F1F1} \raisebox{-1pt}{\small 1.25}} &  \\
\bottomrule
\end{tabularx}
\captionsetup{skip=2pt}
\caption{N = 3}
\end{subtable}
\hfill
\begin{subtable}[h]{0.05\linewidth}
\centering
\includegraphics[height=70pt]{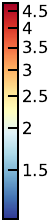}
\vfill \vspace{-0.5em}
\end{subtable}
\hfill
\caption{Median values of the validation loss (x10) on the snapshot dataset.}
\label{tab:loss_median_snap}
\vspace{-0.5em}\end{table*}

%% file: table_median_threshratio.tex
\begin{table*}[p]
\centering
\begin{subtable}[t]{0.29\textwidth}
\centering
\small \begin{tabularx}{\linewidth}{{lXXXX}}
\toprule
\makebox[25pt][r]{$L$} & \makebox[0.5\linewidth][c]{4096} & \makebox[0.8\linewidth][c]{8192} & \makebox[1.1\linewidth][c]{16384} & \makebox[1.4\linewidth][c]{32768} \\
$B$ \textbackslash~ {\color{gray} $L_c$} & \makebox[0.5\linewidth][c]{\color{gray}4096} & \makebox[0.8\linewidth][c]{\color{gray}8192} & \makebox[1.1\linewidth][c]{\color{gray}16384} & \makebox[1.4\linewidth][c]{\color{gray}32768} \\
\midrule
8 & \makebox[\linewidth]{{\cellcolor[HTML]{A90426}} \color[HTML]{F1F1F1} \raisebox{-1pt}{\small 6.44}} & \makebox[\linewidth]{{\cellcolor[HTML]{E44C34}} \color[HTML]{F1F1F1} \raisebox{-1pt}{\small 5.20}} & \makebox[\linewidth]{{\cellcolor[HTML]{F99355}} \color[HTML]{000000} \raisebox{-1pt}{\small 4.37}} & \makebox[\linewidth]{{\cellcolor[HTML]{FED889}} \color[HTML]{000000} \raisebox{-1pt}{\small 3.60}} \\
16 & \makebox[\linewidth]{{\cellcolor[HTML]{A50026}} \color[HTML]{F1F1F1} \raisebox{-1pt}{\small 6.54}} & \makebox[\linewidth]{{\cellcolor[HTML]{F26841}} \color[HTML]{F1F1F1} \raisebox{-1pt}{\small 4.83}} & \makebox[\linewidth]{{\cellcolor[HTML]{FDB164}} \color[HTML]{000000} \raisebox{-1pt}{\small 4.05}} & \makebox[\linewidth]{{\cellcolor[HTML]{E9F6E8}} \color[HTML]{000000} \raisebox{-1pt}{\small 2.68}} \\
32 & \makebox[\linewidth]{{\cellcolor[HTML]{C41E27}} \color[HTML]{F1F1F1} \raisebox{-1pt}{\small 5.91}} & \makebox[\linewidth]{{\cellcolor[HTML]{F36B42}} \color[HTML]{F1F1F1} \raisebox{-1pt}{\small 4.80}} & \makebox[\linewidth]{{\cellcolor[HTML]{EDF8DF}} \color[HTML]{000000} \raisebox{-1pt}{\small 2.75}} & \makebox[\linewidth]{{\cellcolor[HTML]{4065AC}} \color[HTML]{F1F1F1} \raisebox{-1pt}{\small 1.55}} \\
64 & \makebox[\linewidth]{{\cellcolor[HTML]{CE2827}} \color[HTML]{F1F1F1} \raisebox{-1pt}{\small 5.73}} & \makebox[\linewidth]{{\cellcolor[HTML]{FDC374}} \color[HTML]{000000} \raisebox{-1pt}{\small 3.83}} & \makebox[\linewidth]{{\cellcolor[HTML]{FEECA2}} \color[HTML]{000000} \raisebox{-1pt}{\small 3.31}} & \makebox[\linewidth]{{\cellcolor[HTML]{3E60AA}} \color[HTML]{F1F1F1} \raisebox{-1pt}{\small 1.53}} \\
128 & \makebox[\linewidth]{{\cellcolor[HTML]{C62027}} \color[HTML]{F1F1F1} \raisebox{-1pt}{\small 5.86}} & \makebox[\linewidth]{{\cellcolor[HTML]{FEE699}} \color[HTML]{000000} \raisebox{-1pt}{\small 3.39}} & \makebox[\linewidth]{{\cellcolor[HTML]{87BDD9}} \color[HTML]{000000} \raisebox{-1pt}{\small 1.99}} &  \\
\bottomrule
\end{tabularx}
\captionsetup{skip=2pt}
\caption{N = 1}
\end{subtable}
\hfill
\begin{subtable}[t]{0.29\textwidth}
\centering
\small \begin{tabularx}{\linewidth}{{lXXXX}}
\toprule
\makebox[25pt][r]{$L$} & \makebox[0.5\linewidth][c]{4096} & \makebox[0.8\linewidth][c]{8192} & \makebox[1.1\linewidth][c]{16384} & \makebox[1.4\linewidth][c]{32768} \\
$B$ \textbackslash~ {\color{gray} $L_c$} & \makebox[0.5\linewidth][c]{\color{gray}8192} & \makebox[0.8\linewidth][c]{\color{gray}16384} & \makebox[1.1\linewidth][c]{\color{gray}32768} & \makebox[1.4\linewidth][c]{\color{gray}65536} \\
\midrule
8 & \makebox[\linewidth]{{\cellcolor[HTML]{DB382B}} \color[HTML]{F1F1F1} \raisebox{-1pt}{\small 5.47}} & \makebox[\linewidth]{{\cellcolor[HTML]{F7844E}} \color[HTML]{F1F1F1} \raisebox{-1pt}{\small 4.53}} & \makebox[\linewidth]{{\cellcolor[HTML]{FECC7E}} \color[HTML]{000000} \raisebox{-1pt}{\small 3.72}} & \makebox[\linewidth]{{\cellcolor[HTML]{FFF3AD}} \color[HTML]{000000} \raisebox{-1pt}{\small 3.18}} \\
16 & \makebox[\linewidth]{{\cellcolor[HTML]{E0422F}} \color[HTML]{F1F1F1} \raisebox{-1pt}{\small 5.33}} & \makebox[\linewidth]{{\cellcolor[HTML]{FCA85E}} \color[HTML]{000000} \raisebox{-1pt}{\small 4.15}} & \makebox[\linewidth]{{\cellcolor[HTML]{FBFDC7}} \color[HTML]{000000} \raisebox{-1pt}{\small 2.93}} & \makebox[\linewidth]{{\cellcolor[HTML]{9FD0E4}} \color[HTML]{000000} \raisebox{-1pt}{\small 2.12}} \\
32 & \makebox[\linewidth]{{\cellcolor[HTML]{F47044}} \color[HTML]{F1F1F1} \raisebox{-1pt}{\small 4.76}} & \makebox[\linewidth]{{\cellcolor[HTML]{FDB567}} \color[HTML]{000000} \raisebox{-1pt}{\small 4.00}} & \makebox[\linewidth]{{\cellcolor[HTML]{D4EDF4}} \color[HTML]{000000} \raisebox{-1pt}{\small 2.47}} & \makebox[\linewidth]{{\cellcolor[HTML]{6399C7}} \color[HTML]{F1F1F1} \raisebox{-1pt}{\small 1.78}} \\
64 & \makebox[\linewidth]{{\cellcolor[HTML]{F99355}} \color[HTML]{000000} \raisebox{-1pt}{\small 4.36}} & \makebox[\linewidth]{{\cellcolor[HTML]{FDBB6D}} \color[HTML]{000000} \raisebox{-1pt}{\small 3.92}} & \makebox[\linewidth]{{\cellcolor[HTML]{81B7D7}} \color[HTML]{000000} \raisebox{-1pt}{\small 1.96}} & \makebox[\linewidth]{{\cellcolor[HTML]{679EC9}} \color[HTML]{F1F1F1} \raisebox{-1pt}{\small 1.81}} \\
128 & \makebox[\linewidth]{{\cellcolor[HTML]{FA9656}} \color[HTML]{000000} \raisebox{-1pt}{\small 4.33}} & \makebox[\linewidth]{{\cellcolor[HTML]{EAF7E6}} \color[HTML]{000000} \raisebox{-1pt}{\small 2.71}} & \makebox[\linewidth]{{\cellcolor[HTML]{85BBD9}} \color[HTML]{000000} \raisebox{-1pt}{\small 1.98}} &  \\
\bottomrule
\end{tabularx}
\captionsetup{skip=2pt}
\caption{N = 2}
\end{subtable}
\hfill
\begin{subtable}[t]{0.29\textwidth}
\centering
\small \begin{tabularx}{\linewidth}{{lXXXX}}
\toprule
\makebox[25pt][r]{$L$} & \makebox[0.5\linewidth][c]{4096} & \makebox[0.8\linewidth][c]{8192} & \makebox[1.1\linewidth][c]{16384} & \makebox[1.4\linewidth][c]{32768} \\
$B$ \textbackslash~ {\color{gray} $L_c$} & \makebox[0.5\linewidth][c]{\color{gray}12288} & \makebox[0.8\linewidth][c]{\color{gray}24576} & \makebox[1.1\linewidth][c]{\color{gray}49152} & \makebox[1.4\linewidth][c]{\color{gray}98304} \\
\midrule
8 & \makebox[\linewidth]{{\cellcolor[HTML]{F26841}} \color[HTML]{F1F1F1} \raisebox{-1pt}{\small 4.83}} & \makebox[\linewidth]{{\cellcolor[HTML]{FECA7C}} \color[HTML]{000000} \raisebox{-1pt}{\small 3.75}} & \makebox[\linewidth]{{\cellcolor[HTML]{FEE699}} \color[HTML]{000000} \raisebox{-1pt}{\small 3.41}} & \makebox[\linewidth]{{\cellcolor[HTML]{DAF0F6}} \color[HTML]{000000} \raisebox{-1pt}{\small 2.52}} \\
16 & \makebox[\linewidth]{{\cellcolor[HTML]{FCA55D}} \color[HTML]{000000} \raisebox{-1pt}{\small 4.17}} & \makebox[\linewidth]{{\cellcolor[HTML]{FED485}} \color[HTML]{000000} \raisebox{-1pt}{\small 3.64}} & \makebox[\linewidth]{{\cellcolor[HTML]{FCFEC5}} \color[HTML]{000000} \raisebox{-1pt}{\small 2.96}} & \makebox[\linewidth]{{\cellcolor[HTML]{70A9CF}} \color[HTML]{F1F1F1} \raisebox{-1pt}{\small 1.87}} \\
32 & \makebox[\linewidth]{{\cellcolor[HTML]{F8864F}} \color[HTML]{F1F1F1} \raisebox{-1pt}{\small 4.49}} & \makebox[\linewidth]{{\cellcolor[HTML]{FEC87A}} \color[HTML]{000000} \raisebox{-1pt}{\small 3.77}} & \makebox[\linewidth]{{\cellcolor[HTML]{DAF0F6}} \color[HTML]{000000} \raisebox{-1pt}{\small 2.52}} & \makebox[\linewidth]{{\cellcolor[HTML]{74ADD1}} \color[HTML]{F1F1F1} \raisebox{-1pt}{\small 1.89}} \\
64 & \makebox[\linewidth]{{\cellcolor[HTML]{FA9857}} \color[HTML]{000000} \raisebox{-1pt}{\small 4.30}} & \makebox[\linewidth]{{\cellcolor[HTML]{FED283}} \color[HTML]{000000} \raisebox{-1pt}{\small 3.65}} & \makebox[\linewidth]{{\cellcolor[HTML]{92C5DE}} \color[HTML]{000000} \raisebox{-1pt}{\small 2.06}} & \makebox[\linewidth]{{\cellcolor[HTML]{313695}} \color[HTML]{F1F1F1} \raisebox{-1pt}{\small 1.38}} \\
128 & \makebox[\linewidth]{{\cellcolor[HTML]{FBA35C}} \color[HTML]{000000} \raisebox{-1pt}{\small 4.20}} & \makebox[\linewidth]{{\cellcolor[HTML]{CFEBF3}} \color[HTML]{000000} \raisebox{-1pt}{\small 2.45}} & \makebox[\linewidth]{{\cellcolor[HTML]{7AB2D4}} \color[HTML]{000000} \raisebox{-1pt}{\small 1.92}} &  \\
\bottomrule
\end{tabularx}
\captionsetup{skip=2pt}
\caption{N = 3}
\end{subtable}
\hfill
\begin{subtable}[h]{0.05\linewidth}
\centering
\includegraphics[height=70pt]{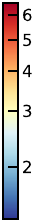}
\vfill \vspace{-0.5em}
\end{subtable}
\hfill
\caption{Median values of the validation loss (x10) on the Threshold-Ratio dataset.}
\label{tab:loss_median_threshratio}
\vspace{-0.5em}\end{table*}

%% file: table_mad_snap.tex
\begin{table*}[p]
\centering
\begin{subtable}[t]{0.29\textwidth}
\centering
\small \begin{tabularx}{\linewidth}{{lXXXX}}
\toprule
\makebox[25pt][r]{$L$} & \makebox[0.5\linewidth][c]{4096} & \makebox[0.8\linewidth][c]{8192} & \makebox[1.1\linewidth][c]{16384} & \makebox[1.4\linewidth][c]{32768} \\
$B$ \textbackslash~ {\color{gray} $L_c$} & \makebox[0.5\linewidth][c]{\color{gray}4096} & \makebox[0.8\linewidth][c]{\color{gray}8192} & \makebox[1.1\linewidth][c]{\color{gray}16384} & \makebox[1.4\linewidth][c]{\color{gray}32768} \\
\midrule
8 & \makebox[\linewidth]{{\cellcolor[HTML]{F16640}} \color[HTML]{F1F1F1} \raisebox{-1pt}{\small 0.79}} & \makebox[\linewidth]{{\cellcolor[HTML]{FCA85E}} \color[HTML]{000000} \raisebox{-1pt}{\small 0.53}} & \makebox[\linewidth]{{\cellcolor[HTML]{F88950}} \color[HTML]{F1F1F1} \raisebox{-1pt}{\small 0.64}} & \makebox[\linewidth]{{\cellcolor[HTML]{FFF0A8}} \color[HTML]{000000} \raisebox{-1pt}{\small 0.27}} \\
16 & \makebox[\linewidth]{{\cellcolor[HTML]{F8864F}} \color[HTML]{F1F1F1} \raisebox{-1pt}{\small 0.65}} & \makebox[\linewidth]{{\cellcolor[HTML]{FCAA5F}} \color[HTML]{000000} \raisebox{-1pt}{\small 0.52}} & \makebox[\linewidth]{{\cellcolor[HTML]{F1FAD9}} \color[HTML]{000000} \raisebox{-1pt}{\small 0.18}} & \makebox[\linewidth]{{\cellcolor[HTML]{ACDAE9}} \color[HTML]{000000} \raisebox{-1pt}{\small 0.10}} \\
32 & \makebox[\linewidth]{{\cellcolor[HTML]{A50026}} \color[HTML]{F1F1F1} \raisebox{-1pt}{\small 1.75}} & \makebox[\linewidth]{{\cellcolor[HTML]{F47044}} \color[HTML]{F1F1F1} \raisebox{-1pt}{\small 0.75}} & \makebox[\linewidth]{{\cellcolor[HTML]{DCF1F7}} \color[HTML]{000000} \raisebox{-1pt}{\small 0.14}} & \makebox[\linewidth]{{\cellcolor[HTML]{426CB0}} \color[HTML]{F1F1F1} \raisebox{-1pt}{\small 0.04}} \\
64 & \makebox[\linewidth]{{\cellcolor[HTML]{FDBD6F}} \color[HTML]{000000} \raisebox{-1pt}{\small 0.44}} & \makebox[\linewidth]{{\cellcolor[HTML]{F98E52}} \color[HTML]{F1F1F1} \raisebox{-1pt}{\small 0.61}} & \makebox[\linewidth]{{\cellcolor[HTML]{C7E7F1}} \color[HTML]{000000} \raisebox{-1pt}{\small 0.12}} & \makebox[\linewidth]{{\cellcolor[HTML]{E9F6E8}} \color[HTML]{000000} \raisebox{-1pt}{\small 0.16}} \\
128 & \makebox[\linewidth]{{\cellcolor[HTML]{E5F5EF}} \color[HTML]{000000} \raisebox{-1pt}{\small 0.15}} & \makebox[\linewidth]{{\cellcolor[HTML]{FDC576}} \color[HTML]{000000} \raisebox{-1pt}{\small 0.42}} & \makebox[\linewidth]{{\cellcolor[HTML]{FEE99D}} \color[HTML]{000000} \raisebox{-1pt}{\small 0.30}} &  \\
\bottomrule
\end{tabularx}
\captionsetup{skip=2pt}
\caption{N = 1}
\end{subtable}
\hfill
\begin{subtable}[t]{0.29\textwidth}
\centering
\small \begin{tabularx}{\linewidth}{{lXXXX}}
\toprule
\makebox[25pt][r]{$L$} & \makebox[0.5\linewidth][c]{4096} & \makebox[0.8\linewidth][c]{8192} & \makebox[1.1\linewidth][c]{16384} & \makebox[1.4\linewidth][c]{32768} \\
$B$ \textbackslash~ {\color{gray} $L_c$} & \makebox[0.5\linewidth][c]{\color{gray}8192} & \makebox[0.8\linewidth][c]{\color{gray}16384} & \makebox[1.1\linewidth][c]{\color{gray}32768} & \makebox[1.4\linewidth][c]{\color{gray}65536} \\
\midrule
8 & \makebox[\linewidth]{{\cellcolor[HTML]{C01A27}} \color[HTML]{F1F1F1} \raisebox{-1pt}{\small 1.37}} & \makebox[\linewidth]{{\cellcolor[HTML]{FFF2AC}} \color[HTML]{000000} \raisebox{-1pt}{\small 0.26}} & \makebox[\linewidth]{{\cellcolor[HTML]{C3E5F0}} \color[HTML]{000000} \raisebox{-1pt}{\small 0.12}} & \makebox[\linewidth]{{\cellcolor[HTML]{5588BE}} \color[HTML]{F1F1F1} \raisebox{-1pt}{\small 0.05}} \\
16 & \makebox[\linewidth]{{\cellcolor[HTML]{FDAD60}} \color[HTML]{000000} \raisebox{-1pt}{\small 0.50}} & \makebox[\linewidth]{{\cellcolor[HTML]{F57245}} \color[HTML]{F1F1F1} \raisebox{-1pt}{\small 0.73}} & \makebox[\linewidth]{{\cellcolor[HTML]{8CC0DB}} \color[HTML]{000000} \raisebox{-1pt}{\small 0.08}} & \makebox[\linewidth]{{\cellcolor[HTML]{94C7DF}} \color[HTML]{000000} \raisebox{-1pt}{\small 0.08}} \\
32 & \makebox[\linewidth]{{\cellcolor[HTML]{FEEFA6}} \color[HTML]{000000} \raisebox{-1pt}{\small 0.28}} & \makebox[\linewidth]{{\cellcolor[HTML]{F98E52}} \color[HTML]{F1F1F1} \raisebox{-1pt}{\small 0.61}} & \makebox[\linewidth]{{\cellcolor[HTML]{F3FBD4}} \color[HTML]{000000} \raisebox{-1pt}{\small 0.19}} & \makebox[\linewidth]{{\cellcolor[HTML]{35429B}} \color[HTML]{F1F1F1} \raisebox{-1pt}{\small 0.03}} \\
64 & \makebox[\linewidth]{{\cellcolor[HTML]{FDC576}} \color[HTML]{000000} \raisebox{-1pt}{\small 0.42}} & \makebox[\linewidth]{{\cellcolor[HTML]{FEEA9F}} \color[HTML]{000000} \raisebox{-1pt}{\small 0.29}} & \makebox[\linewidth]{{\cellcolor[HTML]{E0F3F8}} \color[HTML]{000000} \raisebox{-1pt}{\small 0.14}} & \makebox[\linewidth]{{\cellcolor[HTML]{5588BE}} \color[HTML]{F1F1F1} \raisebox{-1pt}{\small 0.05}} \\
128 & \makebox[\linewidth]{{\cellcolor[HTML]{FECA7C}} \color[HTML]{000000} \raisebox{-1pt}{\small 0.40}} & \makebox[\linewidth]{{\cellcolor[HTML]{FEEA9F}} \color[HTML]{000000} \raisebox{-1pt}{\small 0.29}} & \makebox[\linewidth]{{\cellcolor[HTML]{313695}} \color[HTML]{F1F1F1} \raisebox{-1pt}{\small 0.03}} &  \\
\bottomrule
\end{tabularx}
\captionsetup{skip=2pt}
\caption{N = 2}
\end{subtable}
\hfill
\begin{subtable}[t]{0.29\textwidth}
\centering
\small \begin{tabularx}{\linewidth}{{lXXXX}}
\toprule
\makebox[25pt][r]{$L$} & \makebox[0.5\linewidth][c]{4096} & \makebox[0.8\linewidth][c]{8192} & \makebox[1.1\linewidth][c]{16384} & \makebox[1.4\linewidth][c]{32768} \\
$B$ \textbackslash~ {\color{gray} $L_c$} & \makebox[0.5\linewidth][c]{\color{gray}12288} & \makebox[0.8\linewidth][c]{\color{gray}24576} & \makebox[1.1\linewidth][c]{\color{gray}49152} & \makebox[1.4\linewidth][c]{\color{gray}98304} \\
\midrule
8 & \makebox[\linewidth]{{\cellcolor[HTML]{FB9D59}} \color[HTML]{000000} \raisebox{-1pt}{\small 0.55}} & \makebox[\linewidth]{{\cellcolor[HTML]{EC5C3B}} \color[HTML]{F1F1F1} \raisebox{-1pt}{\small 0.85}} & \makebox[\linewidth]{{\cellcolor[HTML]{ECF8E2}} \color[HTML]{000000} \raisebox{-1pt}{\small 0.17}} & \makebox[\linewidth]{{\cellcolor[HTML]{8ABEDA}} \color[HTML]{000000} \raisebox{-1pt}{\small 0.08}} \\
16 & \makebox[\linewidth]{{\cellcolor[HTML]{FDC576}} \color[HTML]{000000} \raisebox{-1pt}{\small 0.41}} & \makebox[\linewidth]{{\cellcolor[HTML]{F0F9DB}} \color[HTML]{000000} \raisebox{-1pt}{\small 0.18}} & \makebox[\linewidth]{{\cellcolor[HTML]{8ABEDA}} \color[HTML]{000000} \raisebox{-1pt}{\small 0.08}} & \makebox[\linewidth]{{\cellcolor[HTML]{313695}} \color[HTML]{F1F1F1} \raisebox{-1pt}{\small 0.03}} \\
32 & \makebox[\linewidth]{{\cellcolor[HTML]{FECA7C}} \color[HTML]{000000} \raisebox{-1pt}{\small 0.40}} & \makebox[\linewidth]{{\cellcolor[HTML]{FED283}} \color[HTML]{000000} \raisebox{-1pt}{\small 0.37}} & \makebox[\linewidth]{{\cellcolor[HTML]{C3E5F0}} \color[HTML]{000000} \raisebox{-1pt}{\small 0.12}} & \makebox[\linewidth]{{\cellcolor[HTML]{70A9CF}} \color[HTML]{F1F1F1} \raisebox{-1pt}{\small 0.06}} \\
64 & \makebox[\linewidth]{{\cellcolor[HTML]{FDFEC2}} \color[HTML]{000000} \raisebox{-1pt}{\small 0.22}} & \makebox[\linewidth]{{\cellcolor[HTML]{D4EDF4}} \color[HTML]{000000} \raisebox{-1pt}{\small 0.13}} & \makebox[\linewidth]{{\cellcolor[HTML]{3F62AB}} \color[HTML]{F1F1F1} \raisebox{-1pt}{\small 0.04}} & \makebox[\linewidth]{{\cellcolor[HTML]{8EC2DC}} \color[HTML]{000000} \raisebox{-1pt}{\small 0.08}} \\
128 & \makebox[\linewidth]{{\cellcolor[HTML]{E6F5ED}} \color[HTML]{000000} \raisebox{-1pt}{\small 0.16}} & \makebox[\linewidth]{{\cellcolor[HTML]{FED687}} \color[HTML]{000000} \raisebox{-1pt}{\small 0.36}} & \makebox[\linewidth]{{\cellcolor[HTML]{70A9CF}} \color[HTML]{F1F1F1} \raisebox{-1pt}{\small 0.06}} &  \\
\bottomrule
\end{tabularx}
\captionsetup{skip=2pt}
\caption{N = 3}
\end{subtable}
\hfill
\begin{subtable}[h]{0.05\linewidth}
\centering
\includegraphics[height=70pt]{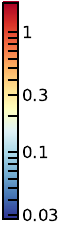}
\vfill \vspace{-0.5em}
\end{subtable}
\hfill
\caption{Median Absolute Deviation (MAD) values of the validation loss (x10) on the snapshot dataset.}
\label{tab:loss_mad_snap}
\vspace{-0.5em}\end{table*}

%% file: table_mad_threshratio.tex
\begin{table*}[p]
\centering
\begin{subtable}[t]{0.29\textwidth}
\centering
\small \begin{tabularx}{\linewidth}{{lXXXX}}
\toprule
\makebox[25pt][r]{$L$} & \makebox[0.5\linewidth][c]{4096} & \makebox[0.8\linewidth][c]{8192} & \makebox[1.1\linewidth][c]{16384} & \makebox[1.4\linewidth][c]{32768} \\
$B$ \textbackslash~ {\color{gray} $L_c$} & \makebox[0.5\linewidth][c]{\color{gray}4096} & \makebox[0.8\linewidth][c]{\color{gray}8192} & \makebox[1.1\linewidth][c]{\color{gray}16384} & \makebox[1.4\linewidth][c]{\color{gray}32768} \\
\midrule
8 & \makebox[\linewidth]{{\cellcolor[HTML]{FDC374}} \color[HTML]{000000} \raisebox{-1pt}{\small 0.79}} & \makebox[\linewidth]{{\cellcolor[HTML]{6DA4CC}} \color[HTML]{F1F1F1} \raisebox{-1pt}{\small 0.24}} & \makebox[\linewidth]{{\cellcolor[HTML]{EAF7E6}} \color[HTML]{000000} \raisebox{-1pt}{\small 0.44}} & \makebox[\linewidth]{{\cellcolor[HTML]{F3FBD4}} \color[HTML]{000000} \raisebox{-1pt}{\small 0.48}} \\
16 & \makebox[\linewidth]{{\cellcolor[HTML]{FDC173}} \color[HTML]{000000} \raisebox{-1pt}{\small 0.80}} & \makebox[\linewidth]{{\cellcolor[HTML]{FED081}} \color[HTML]{000000} \raisebox{-1pt}{\small 0.74}} & \makebox[\linewidth]{{\cellcolor[HTML]{FDC173}} \color[HTML]{000000} \raisebox{-1pt}{\small 0.80}} & \makebox[\linewidth]{{\cellcolor[HTML]{FFF0A8}} \color[HTML]{000000} \raisebox{-1pt}{\small 0.59}} \\
32 & \makebox[\linewidth]{{\cellcolor[HTML]{FA9857}} \color[HTML]{000000} \raisebox{-1pt}{\small 0.96}} & \makebox[\linewidth]{{\cellcolor[HTML]{FFFBB9}} \color[HTML]{000000} \raisebox{-1pt}{\small 0.55}} & \makebox[\linewidth]{{\cellcolor[HTML]{F8864F}} \color[HTML]{F1F1F1} \raisebox{-1pt}{\small 1.03}} & \makebox[\linewidth]{{\cellcolor[HTML]{8ABEDA}} \color[HTML]{000000} \raisebox{-1pt}{\small 0.27}} \\
64 & \makebox[\linewidth]{{\cellcolor[HTML]{F57245}} \color[HTML]{F1F1F1} \raisebox{-1pt}{\small 1.11}} & \makebox[\linewidth]{{\cellcolor[HTML]{C41E27}} \color[HTML]{F1F1F1} \raisebox{-1pt}{\small 1.61}} & \makebox[\linewidth]{{\cellcolor[HTML]{FEDC8C}} \color[HTML]{000000} \raisebox{-1pt}{\small 0.69}} & \makebox[\linewidth]{{\cellcolor[HTML]{9DCEE3}} \color[HTML]{000000} \raisebox{-1pt}{\small 0.30}} \\
128 & \makebox[\linewidth]{{\cellcolor[HTML]{A50026}} \color[HTML]{F1F1F1} \raisebox{-1pt}{\small 1.90}} & \makebox[\linewidth]{{\cellcolor[HTML]{E24731}} \color[HTML]{F1F1F1} \raisebox{-1pt}{\small 1.33}} & \makebox[\linewidth]{{\cellcolor[HTML]{F8FCCB}} \color[HTML]{000000} \raisebox{-1pt}{\small 0.50}} &  \\
\bottomrule
\end{tabularx}
\captionsetup{skip=2pt}
\caption{N = 1}
\end{subtable}
\hfill
\begin{subtable}[t]{0.29\textwidth}
\centering
\small \begin{tabularx}{\linewidth}{{lXXXX}}
\toprule
\makebox[25pt][r]{$L$} & \makebox[0.5\linewidth][c]{4096} & \makebox[0.8\linewidth][c]{8192} & \makebox[1.1\linewidth][c]{16384} & \makebox[1.4\linewidth][c]{32768} \\
$B$ \textbackslash~ {\color{gray} $L_c$} & \makebox[0.5\linewidth][c]{\color{gray}8192} & \makebox[0.8\linewidth][c]{\color{gray}16384} & \makebox[1.1\linewidth][c]{\color{gray}32768} & \makebox[1.4\linewidth][c]{\color{gray}65536} \\
\midrule
8 & \makebox[\linewidth]{{\cellcolor[HTML]{E9F6E8}} \color[HTML]{000000} \raisebox{-1pt}{\small 0.44}} & \makebox[\linewidth]{{\cellcolor[HTML]{FFF2AC}} \color[HTML]{000000} \raisebox{-1pt}{\small 0.59}} & \makebox[\linewidth]{{\cellcolor[HTML]{E2F4F4}} \color[HTML]{000000} \raisebox{-1pt}{\small 0.42}} & \makebox[\linewidth]{{\cellcolor[HTML]{8ABEDA}} \color[HTML]{000000} \raisebox{-1pt}{\small 0.27}} \\
16 & \makebox[\linewidth]{{\cellcolor[HTML]{FEEBA1}} \color[HTML]{000000} \raisebox{-1pt}{\small 0.62}} & \makebox[\linewidth]{{\cellcolor[HTML]{FBFDC7}} \color[HTML]{000000} \raisebox{-1pt}{\small 0.51}} & \makebox[\linewidth]{{\cellcolor[HTML]{FCFEC5}} \color[HTML]{000000} \raisebox{-1pt}{\small 0.51}} & \makebox[\linewidth]{{\cellcolor[HTML]{D1ECF4}} \color[HTML]{000000} \raisebox{-1pt}{\small 0.38}} \\
32 & \makebox[\linewidth]{{\cellcolor[HTML]{FDFEC2}} \color[HTML]{000000} \raisebox{-1pt}{\small 0.52}} & \makebox[\linewidth]{{\cellcolor[HTML]{FCFEC5}} \color[HTML]{000000} \raisebox{-1pt}{\small 0.51}} & \makebox[\linewidth]{{\cellcolor[HTML]{FFFCBA}} \color[HTML]{000000} \raisebox{-1pt}{\small 0.54}} & \makebox[\linewidth]{{\cellcolor[HTML]{F7FCCE}} \color[HTML]{000000} \raisebox{-1pt}{\small 0.49}} \\
64 & \makebox[\linewidth]{{\cellcolor[HTML]{F1FAD9}} \color[HTML]{000000} \raisebox{-1pt}{\small 0.47}} & \makebox[\linewidth]{{\cellcolor[HTML]{FFF7B3}} \color[HTML]{000000} \raisebox{-1pt}{\small 0.56}} & \makebox[\linewidth]{{\cellcolor[HTML]{92C5DE}} \color[HTML]{000000} \raisebox{-1pt}{\small 0.28}} & \makebox[\linewidth]{{\cellcolor[HTML]{6DA4CC}} \color[HTML]{F1F1F1} \raisebox{-1pt}{\small 0.24}} \\
128 & \makebox[\linewidth]{{\cellcolor[HTML]{FFFCBA}} \color[HTML]{000000} \raisebox{-1pt}{\small 0.54}} & \makebox[\linewidth]{{\cellcolor[HTML]{FFF8B5}} \color[HTML]{000000} \raisebox{-1pt}{\small 0.56}} & \makebox[\linewidth]{{\cellcolor[HTML]{74ADD1}} \color[HTML]{F1F1F1} \raisebox{-1pt}{\small 0.25}} &  \\
\bottomrule
\end{tabularx}
\captionsetup{skip=2pt}
\caption{N = 2}
\end{subtable}
\hfill
\begin{subtable}[t]{0.29\textwidth}
\centering
\small \begin{tabularx}{\linewidth}{{lXXXX}}
\toprule
\makebox[25pt][r]{$L$} & \makebox[0.5\linewidth][c]{4096} & \makebox[0.8\linewidth][c]{8192} & \makebox[1.1\linewidth][c]{16384} & \makebox[1.4\linewidth][c]{32768} \\
$B$ \textbackslash~ {\color{gray} $L_c$} & \makebox[0.5\linewidth][c]{\color{gray}12288} & \makebox[0.8\linewidth][c]{\color{gray}24576} & \makebox[1.1\linewidth][c]{\color{gray}49152} & \makebox[1.4\linewidth][c]{\color{gray}98304} \\
\midrule
8 & \makebox[\linewidth]{{\cellcolor[HTML]{DEF2F7}} \color[HTML]{000000} \raisebox{-1pt}{\small 0.41}} & \makebox[\linewidth]{{\cellcolor[HTML]{87BDD9}} \color[HTML]{000000} \raisebox{-1pt}{\small 0.27}} & \makebox[\linewidth]{{\cellcolor[HTML]{FED889}} \color[HTML]{000000} \raisebox{-1pt}{\small 0.71}} & \makebox[\linewidth]{{\cellcolor[HTML]{E0F3F8}} \color[HTML]{000000} \raisebox{-1pt}{\small 0.41}} \\
16 & \makebox[\linewidth]{{\cellcolor[HTML]{CBE9F2}} \color[HTML]{000000} \raisebox{-1pt}{\small 0.37}} & \makebox[\linewidth]{{\cellcolor[HTML]{A1D1E5}} \color[HTML]{000000} \raisebox{-1pt}{\small 0.30}} & \makebox[\linewidth]{{\cellcolor[HTML]{FFF8B5}} \color[HTML]{000000} \raisebox{-1pt}{\small 0.55}} & \makebox[\linewidth]{{\cellcolor[HTML]{588CC0}} \color[HTML]{F1F1F1} \raisebox{-1pt}{\small 0.21}} \\
32 & \makebox[\linewidth]{{\cellcolor[HTML]{A1D1E5}} \color[HTML]{000000} \raisebox{-1pt}{\small 0.30}} & \makebox[\linewidth]{{\cellcolor[HTML]{D1ECF4}} \color[HTML]{000000} \raisebox{-1pt}{\small 0.38}} & \makebox[\linewidth]{{\cellcolor[HTML]{85BBD9}} \color[HTML]{000000} \raisebox{-1pt}{\small 0.26}} & \makebox[\linewidth]{{\cellcolor[HTML]{C1E4EF}} \color[HTML]{000000} \raisebox{-1pt}{\small 0.35}} \\
64 & \makebox[\linewidth]{{\cellcolor[HTML]{F1FAD9}} \color[HTML]{000000} \raisebox{-1pt}{\small 0.47}} & \makebox[\linewidth]{{\cellcolor[HTML]{99CAE1}} \color[HTML]{000000} \raisebox{-1pt}{\small 0.29}} & \makebox[\linewidth]{{\cellcolor[HTML]{BFE3EF}} \color[HTML]{000000} \raisebox{-1pt}{\small 0.35}} & \makebox[\linewidth]{{\cellcolor[HTML]{313695}} \color[HTML]{F1F1F1} \raisebox{-1pt}{\small 0.15}} \\
128 & \makebox[\linewidth]{{\cellcolor[HTML]{E44C34}} \color[HTML]{F1F1F1} \raisebox{-1pt}{\small 1.31}} & \makebox[\linewidth]{{\cellcolor[HTML]{D8EFF6}} \color[HTML]{000000} \raisebox{-1pt}{\small 0.39}} & \makebox[\linewidth]{{\cellcolor[HTML]{B4DEEC}} \color[HTML]{000000} \raisebox{-1pt}{\small 0.33}} &  \\
\bottomrule
\end{tabularx}
\captionsetup{skip=2pt}
\caption{N = 3}
\end{subtable}
\hfill
\begin{subtable}[h]{0.05\linewidth}
\centering
\includegraphics[height=70pt]{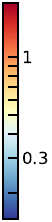}
\vfill \vspace{-0.5em}
\end{subtable}
\hfill
\caption{Median Absolute Deviation (MAD) values of the validation loss (x10) on the Threshold-Ratio dataset.}
\label{tab:loss_mad_threshratio}
\vspace{-0.5em}\end{table*}

%% file: table_time_median_snap.tex
\begin{table*}[p]
\centering
\begin{subtable}[t]{0.29\textwidth}
\centering
\small \begin{tabularx}{\linewidth}{{lXXXX}}
\toprule
\makebox[25pt][r]{$L$} & \makebox[0.5\linewidth][c]{4096} & \makebox[0.8\linewidth][c]{8192} & \makebox[1.1\linewidth][c]{16384} & \makebox[1.4\linewidth][c]{32768} \\
$B$ \textbackslash~ {\color{gray} $L_c$} & \makebox[0.5\linewidth][c]{\color{gray}4096} & \makebox[0.8\linewidth][c]{\color{gray}8192} & \makebox[1.1\linewidth][c]{\color{gray}16384} & \makebox[1.4\linewidth][c]{\color{gray}32768} \\
\midrule
8 & \makebox[\linewidth]{{\cellcolor[HTML]{4676B5}} \color[HTML]{F1F1F1} \raisebox{-1pt}{\small 0.40}} & \makebox[\linewidth]{{\cellcolor[HTML]{313695}} \color[HTML]{F1F1F1} \raisebox{-1pt}{\small 0.28}} & \makebox[\linewidth]{{\cellcolor[HTML]{9DCEE3}} \color[HTML]{000000} \raisebox{-1pt}{\small 0.74}} & \makebox[\linewidth]{{\cellcolor[HTML]{E0F3F8}} \color[HTML]{000000} \raisebox{-1pt}{\small 1.16}} \\
16 & \makebox[\linewidth]{{\cellcolor[HTML]{578ABF}} \color[HTML]{F1F1F1} \raisebox{-1pt}{\small 0.46}} & \makebox[\linewidth]{{\cellcolor[HTML]{7DB4D5}} \color[HTML]{000000} \raisebox{-1pt}{\small 0.60}} & \makebox[\linewidth]{{\cellcolor[HTML]{CDEAF3}} \color[HTML]{000000} \raisebox{-1pt}{\small 1.03}} & \makebox[\linewidth]{{\cellcolor[HTML]{FDC778}} \color[HTML]{000000} \raisebox{-1pt}{\small 2.83}} \\
32 & \makebox[\linewidth]{{\cellcolor[HTML]{6297C6}} \color[HTML]{F1F1F1} \raisebox{-1pt}{\small 0.50}} & \makebox[\linewidth]{{\cellcolor[HTML]{ACDAE9}} \color[HTML]{000000} \raisebox{-1pt}{\small 0.82}} & \makebox[\linewidth]{{\cellcolor[HTML]{FDC778}} \color[HTML]{000000} \raisebox{-1pt}{\small 2.84}} & \makebox[\linewidth]{{\cellcolor[HTML]{FDB96B}} \color[HTML]{000000} \raisebox{-1pt}{\small 3.11}} \\
64 & \makebox[\linewidth]{{\cellcolor[HTML]{E4F4F1}} \color[HTML]{000000} \raisebox{-1pt}{\small 1.20}} & \makebox[\linewidth]{{\cellcolor[HTML]{FFF0A8}} \color[HTML]{000000} \raisebox{-1pt}{\small 1.98}} & \makebox[\linewidth]{{\cellcolor[HTML]{C01A27}} \color[HTML]{F1F1F1} \raisebox{-1pt}{\small 8.05}} & \makebox[\linewidth]{{\cellcolor[HTML]{A50026}} \color[HTML]{F1F1F1} \raisebox{-1pt}{\small 9.82}} \\
128 & \makebox[\linewidth]{{\cellcolor[HTML]{F99153}} \color[HTML]{000000} \raisebox{-1pt}{\small 3.95}} & \makebox[\linewidth]{{\cellcolor[HTML]{FECC7E}} \color[HTML]{000000} \raisebox{-1pt}{\small 2.71}} & \makebox[\linewidth]{{\cellcolor[HTML]{F67F4B}} \color[HTML]{F1F1F1} \raisebox{-1pt}{\small 4.36}} &  \\
\bottomrule
\end{tabularx}
\captionsetup{skip=2pt}
\caption{N = 1}
\end{subtable}
\hfill
\begin{subtable}[t]{0.29\textwidth}
\centering
\small \begin{tabularx}{\linewidth}{{lXXXX}}
\toprule
\makebox[25pt][r]{$L$} & \makebox[0.5\linewidth][c]{4096} & \makebox[0.8\linewidth][c]{8192} & \makebox[1.1\linewidth][c]{16384} & \makebox[1.4\linewidth][c]{32768} \\
$B$ \textbackslash~ {\color{gray} $L_c$} & \makebox[0.5\linewidth][c]{\color{gray}8192} & \makebox[0.8\linewidth][c]{\color{gray}16384} & \makebox[1.1\linewidth][c]{\color{gray}32768} & \makebox[1.4\linewidth][c]{\color{gray}65536} \\
\midrule
8 & \makebox[\linewidth]{{\cellcolor[HTML]{69A0CA}} \color[HTML]{F1F1F1} \raisebox{-1pt}{\small 0.53}} & \makebox[\linewidth]{{\cellcolor[HTML]{588CC0}} \color[HTML]{F1F1F1} \raisebox{-1pt}{\small 0.47}} & \makebox[\linewidth]{{\cellcolor[HTML]{85BBD9}} \color[HTML]{000000} \raisebox{-1pt}{\small 0.64}} & \makebox[\linewidth]{{\cellcolor[HTML]{ACDAE9}} \color[HTML]{000000} \raisebox{-1pt}{\small 0.82}} \\
16 & \makebox[\linewidth]{{\cellcolor[HTML]{5588BE}} \color[HTML]{F1F1F1} \raisebox{-1pt}{\small 0.45}} & \makebox[\linewidth]{{\cellcolor[HTML]{578ABF}} \color[HTML]{F1F1F1} \raisebox{-1pt}{\small 0.46}} & \makebox[\linewidth]{{\cellcolor[HTML]{B4DEEC}} \color[HTML]{000000} \raisebox{-1pt}{\small 0.87}} & \makebox[\linewidth]{{\cellcolor[HTML]{DEF2F7}} \color[HTML]{000000} \raisebox{-1pt}{\small 1.15}} \\
32 & \makebox[\linewidth]{{\cellcolor[HTML]{B6DFEC}} \color[HTML]{000000} \raisebox{-1pt}{\small 0.87}} & \makebox[\linewidth]{{\cellcolor[HTML]{BBE1ED}} \color[HTML]{000000} \raisebox{-1pt}{\small 0.91}} & \makebox[\linewidth]{{\cellcolor[HTML]{E7F6EB}} \color[HTML]{000000} \raisebox{-1pt}{\small 1.27}} & \makebox[\linewidth]{{\cellcolor[HTML]{FDC778}} \color[HTML]{000000} \raisebox{-1pt}{\small 2.82}} \\
64 & \makebox[\linewidth]{{\cellcolor[HTML]{FFF0A8}} \color[HTML]{000000} \raisebox{-1pt}{\small 1.97}} & \makebox[\linewidth]{{\cellcolor[HTML]{EBF7E4}} \color[HTML]{000000} \raisebox{-1pt}{\small 1.32}} & \makebox[\linewidth]{{\cellcolor[HTML]{FFF0A8}} \color[HTML]{000000} \raisebox{-1pt}{\small 1.96}} & \makebox[\linewidth]{{\cellcolor[HTML]{E34933}} \color[HTML]{F1F1F1} \raisebox{-1pt}{\small 5.94}} \\
128 & \makebox[\linewidth]{{\cellcolor[HTML]{FDFEC2}} \color[HTML]{000000} \raisebox{-1pt}{\small 1.61}} & \makebox[\linewidth]{{\cellcolor[HTML]{FA9857}} \color[HTML]{000000} \raisebox{-1pt}{\small 3.79}} & \makebox[\linewidth]{{\cellcolor[HTML]{FDAD60}} \color[HTML]{000000} \raisebox{-1pt}{\small 3.40}} &  \\
\bottomrule
\end{tabularx}
\captionsetup{skip=2pt}
\caption{N = 2}
\end{subtable}
\hfill
\begin{subtable}[t]{0.29\textwidth}
\centering
\small \begin{tabularx}{\linewidth}{{lXXXX}}
\toprule
\makebox[25pt][r]{$L$} & \makebox[0.5\linewidth][c]{4096} & \makebox[0.8\linewidth][c]{8192} & \makebox[1.1\linewidth][c]{16384} & \makebox[1.4\linewidth][c]{32768} \\
$B$ \textbackslash~ {\color{gray} $L_c$} & \makebox[0.5\linewidth][c]{\color{gray}12288} & \makebox[0.8\linewidth][c]{\color{gray}24576} & \makebox[1.1\linewidth][c]{\color{gray}49152} & \makebox[1.4\linewidth][c]{\color{gray}98304} \\
\midrule
8 & \makebox[\linewidth]{{\cellcolor[HTML]{36459C}} \color[HTML]{F1F1F1} \raisebox{-1pt}{\small 0.31}} & \makebox[\linewidth]{{\cellcolor[HTML]{7FB6D6}} \color[HTML]{000000} \raisebox{-1pt}{\small 0.61}} & \makebox[\linewidth]{{\cellcolor[HTML]{6BA2CB}} \color[HTML]{F1F1F1} \raisebox{-1pt}{\small 0.53}} & \makebox[\linewidth]{{\cellcolor[HTML]{E7F6EB}} \color[HTML]{000000} \raisebox{-1pt}{\small 1.26}} \\
16 & \makebox[\linewidth]{{\cellcolor[HTML]{5183BB}} \color[HTML]{F1F1F1} \raisebox{-1pt}{\small 0.44}} & \makebox[\linewidth]{{\cellcolor[HTML]{74ADD1}} \color[HTML]{F1F1F1} \raisebox{-1pt}{\small 0.57}} & \makebox[\linewidth]{{\cellcolor[HTML]{D1ECF4}} \color[HTML]{000000} \raisebox{-1pt}{\small 1.06}} & \makebox[\linewidth]{{\cellcolor[HTML]{DCF1F7}} \color[HTML]{000000} \raisebox{-1pt}{\small 1.13}} \\
32 & \makebox[\linewidth]{{\cellcolor[HTML]{6297C6}} \color[HTML]{F1F1F1} \raisebox{-1pt}{\small 0.49}} & \makebox[\linewidth]{{\cellcolor[HTML]{A3D3E6}} \color[HTML]{000000} \raisebox{-1pt}{\small 0.78}} & \makebox[\linewidth]{{\cellcolor[HTML]{DCF1F7}} \color[HTML]{000000} \raisebox{-1pt}{\small 1.14}} & \makebox[\linewidth]{{\cellcolor[HTML]{F0F9DB}} \color[HTML]{000000} \raisebox{-1pt}{\small 1.39}} \\
64 & \makebox[\linewidth]{{\cellcolor[HTML]{B6DFEC}} \color[HTML]{000000} \raisebox{-1pt}{\small 0.88}} & \makebox[\linewidth]{{\cellcolor[HTML]{FFF5AF}} \color[HTML]{000000} \raisebox{-1pt}{\small 1.87}} & \makebox[\linewidth]{{\cellcolor[HTML]{FBA05B}} \color[HTML]{000000} \raisebox{-1pt}{\small 3.63}} & \makebox[\linewidth]{{\cellcolor[HTML]{F57748}} \color[HTML]{F1F1F1} \raisebox{-1pt}{\small 4.51}} \\
128 & \makebox[\linewidth]{{\cellcolor[HTML]{FEE699}} \color[HTML]{000000} \raisebox{-1pt}{\small 2.20}} & \makebox[\linewidth]{{\cellcolor[HTML]{FCAA5F}} \color[HTML]{000000} \raisebox{-1pt}{\small 3.44}} & \makebox[\linewidth]{{\cellcolor[HTML]{C01A27}} \color[HTML]{F1F1F1} \raisebox{-1pt}{\small 8.02}} &  \\
\bottomrule
\end{tabularx}
\captionsetup{skip=2pt}
\caption{N = 3}
\end{subtable}
\hfill
\begin{subtable}[h]{0.05\linewidth}
\centering
\includegraphics[height=70pt]{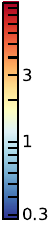}
\vfill \vspace{-0.5em}
\end{subtable}
\hfill
\caption{Median values of the training time (in hours) on the snapshot dataset.}
\label{tab:time_median_snap}
\vspace{-0.5em}\end{table*}

%% file: table_time_median_threshratio.tex
\begin{table*}[p]
\centering
\begin{subtable}[t]{0.29\textwidth}
\centering
\small \begin{tabularx}{\linewidth}{{lXXXX}}
\toprule
\makebox[25pt][r]{$L$} & \makebox[0.5\linewidth][c]{4096} & \makebox[0.8\linewidth][c]{8192} & \makebox[1.1\linewidth][c]{16384} & \makebox[1.4\linewidth][c]{32768} \\
$B$ \textbackslash~ {\color{gray} $L_c$} & \makebox[0.5\linewidth][c]{\color{gray}4096} & \makebox[0.8\linewidth][c]{\color{gray}8192} & \makebox[1.1\linewidth][c]{\color{gray}16384} & \makebox[1.4\linewidth][c]{\color{gray}32768} \\
\midrule
8 & \makebox[\linewidth]{{\cellcolor[HTML]{313695}} \color[HTML]{F1F1F1} \raisebox{-1pt}{\small 0.36}} & \makebox[\linewidth]{{\cellcolor[HTML]{4B7DB8}} \color[HTML]{F1F1F1} \raisebox{-1pt}{\small 0.52}} & \makebox[\linewidth]{{\cellcolor[HTML]{78B0D3}} \color[HTML]{F1F1F1} \raisebox{-1pt}{\small 0.73}} & \makebox[\linewidth]{{\cellcolor[HTML]{3A54A4}} \color[HTML]{F1F1F1} \raisebox{-1pt}{\small 0.42}} \\
16 & \makebox[\linewidth]{{\cellcolor[HTML]{3E60AA}} \color[HTML]{F1F1F1} \raisebox{-1pt}{\small 0.45}} & \makebox[\linewidth]{{\cellcolor[HTML]{578ABF}} \color[HTML]{F1F1F1} \raisebox{-1pt}{\small 0.57}} & \makebox[\linewidth]{{\cellcolor[HTML]{81B7D7}} \color[HTML]{000000} \raisebox{-1pt}{\small 0.77}} & \makebox[\linewidth]{{\cellcolor[HTML]{9BCCE2}} \color[HTML]{000000} \raisebox{-1pt}{\small 0.90}} \\
32 & \makebox[\linewidth]{{\cellcolor[HTML]{99CAE1}} \color[HTML]{000000} \raisebox{-1pt}{\small 0.89}} & \makebox[\linewidth]{{\cellcolor[HTML]{C1E4EF}} \color[HTML]{000000} \raisebox{-1pt}{\small 1.15}} & \makebox[\linewidth]{{\cellcolor[HTML]{FFF8B5}} \color[HTML]{000000} \raisebox{-1pt}{\small 2.13}} & \makebox[\linewidth]{{\cellcolor[HTML]{FECA7C}} \color[HTML]{000000} \raisebox{-1pt}{\small 3.22}} \\
64 & \makebox[\linewidth]{{\cellcolor[HTML]{E6F5ED}} \color[HTML]{000000} \raisebox{-1pt}{\small 1.50}} & \makebox[\linewidth]{{\cellcolor[HTML]{F0F9DB}} \color[HTML]{000000} \raisebox{-1pt}{\small 1.67}} & \makebox[\linewidth]{{\cellcolor[HTML]{FDB164}} \color[HTML]{000000} \raisebox{-1pt}{\small 3.88}} & \makebox[\linewidth]{{\cellcolor[HTML]{B71126}} \color[HTML]{F1F1F1} \raisebox{-1pt}{\small 9.74}} \\
128 & \makebox[\linewidth]{{\cellcolor[HTML]{FDFEC2}} \color[HTML]{000000} \raisebox{-1pt}{\small 1.94}} & \makebox[\linewidth]{{\cellcolor[HTML]{F88950}} \color[HTML]{F1F1F1} \raisebox{-1pt}{\small 4.76}} & \makebox[\linewidth]{{\cellcolor[HTML]{A50026}} \color[HTML]{F1F1F1} \raisebox{-1pt}{\small 11.12}} &  \\
\bottomrule
\end{tabularx}
\captionsetup{skip=2pt}
\caption{N = 1}
\end{subtable}
\hfill
\begin{subtable}[t]{0.29\textwidth}
\centering
\small \begin{tabularx}{\linewidth}{{lXXXX}}
\toprule
\makebox[25pt][r]{$L$} & \makebox[0.5\linewidth][c]{4096} & \makebox[0.8\linewidth][c]{8192} & \makebox[1.1\linewidth][c]{16384} & \makebox[1.4\linewidth][c]{32768} \\
$B$ \textbackslash~ {\color{gray} $L_c$} & \makebox[0.5\linewidth][c]{\color{gray}8192} & \makebox[0.8\linewidth][c]{\color{gray}16384} & \makebox[1.1\linewidth][c]{\color{gray}32768} & \makebox[1.4\linewidth][c]{\color{gray}65536} \\
\midrule
8 & \makebox[\linewidth]{{\cellcolor[HTML]{384CA0}} \color[HTML]{F1F1F1} \raisebox{-1pt}{\small 0.40}} & \makebox[\linewidth]{{\cellcolor[HTML]{6297C6}} \color[HTML]{F1F1F1} \raisebox{-1pt}{\small 0.62}} & \makebox[\linewidth]{{\cellcolor[HTML]{4065AC}} \color[HTML]{F1F1F1} \raisebox{-1pt}{\small 0.46}} & \makebox[\linewidth]{{\cellcolor[HTML]{313695}} \color[HTML]{F1F1F1} \raisebox{-1pt}{\small 0.35}} \\
16 & \makebox[\linewidth]{{\cellcolor[HTML]{3C59A6}} \color[HTML]{F1F1F1} \raisebox{-1pt}{\small 0.43}} & \makebox[\linewidth]{{\cellcolor[HTML]{3A54A4}} \color[HTML]{F1F1F1} \raisebox{-1pt}{\small 0.42}} & \makebox[\linewidth]{{\cellcolor[HTML]{97C9E0}} \color[HTML]{000000} \raisebox{-1pt}{\small 0.88}} & \makebox[\linewidth]{{\cellcolor[HTML]{BBE1ED}} \color[HTML]{000000} \raisebox{-1pt}{\small 1.10}} \\
32 & \makebox[\linewidth]{{\cellcolor[HTML]{588CC0}} \color[HTML]{F1F1F1} \raisebox{-1pt}{\small 0.58}} & \makebox[\linewidth]{{\cellcolor[HTML]{AAD8E9}} \color[HTML]{000000} \raisebox{-1pt}{\small 0.99}} & \makebox[\linewidth]{{\cellcolor[HTML]{DAF0F6}} \color[HTML]{000000} \raisebox{-1pt}{\small 1.35}} & \makebox[\linewidth]{{\cellcolor[HTML]{FBFDC7}} \color[HTML]{000000} \raisebox{-1pt}{\small 1.89}} \\
64 & \makebox[\linewidth]{{\cellcolor[HTML]{C7E7F1}} \color[HTML]{000000} \raisebox{-1pt}{\small 1.20}} & \makebox[\linewidth]{{\cellcolor[HTML]{A3D3E6}} \color[HTML]{000000} \raisebox{-1pt}{\small 0.95}} & \makebox[\linewidth]{{\cellcolor[HTML]{FEEBA1}} \color[HTML]{000000} \raisebox{-1pt}{\small 2.47}} & \makebox[\linewidth]{{\cellcolor[HTML]{FDAF62}} \color[HTML]{000000} \raisebox{-1pt}{\small 3.93}} \\
128 & \makebox[\linewidth]{{\cellcolor[HTML]{FFF8B5}} \color[HTML]{000000} \raisebox{-1pt}{\small 2.13}} & \makebox[\linewidth]{{\cellcolor[HTML]{FEE699}} \color[HTML]{000000} \raisebox{-1pt}{\small 2.62}} & \makebox[\linewidth]{{\cellcolor[HTML]{EC5C3B}} \color[HTML]{F1F1F1} \raisebox{-1pt}{\small 6.14}} &  \\
\bottomrule
\end{tabularx}
\captionsetup{skip=2pt}
\caption{N = 2}
\end{subtable}
\hfill
\begin{subtable}[t]{0.29\textwidth}
\centering
\small \begin{tabularx}{\linewidth}{{lXXXX}}
\toprule
\makebox[25pt][r]{$L$} & \makebox[0.5\linewidth][c]{4096} & \makebox[0.8\linewidth][c]{8192} & \makebox[1.1\linewidth][c]{16384} & \makebox[1.4\linewidth][c]{32768} \\
$B$ \textbackslash~ {\color{gray} $L_c$} & \makebox[0.5\linewidth][c]{\color{gray}12288} & \makebox[0.8\linewidth][c]{\color{gray}24576} & \makebox[1.1\linewidth][c]{\color{gray}49152} & \makebox[1.4\linewidth][c]{\color{gray}98304} \\
\midrule
8 & \makebox[\linewidth]{{\cellcolor[HTML]{5183BB}} \color[HTML]{F1F1F1} \raisebox{-1pt}{\small 0.55}} & \makebox[\linewidth]{{\cellcolor[HTML]{4B7DB8}} \color[HTML]{F1F1F1} \raisebox{-1pt}{\small 0.52}} & \makebox[\linewidth]{{\cellcolor[HTML]{6EA6CE}} \color[HTML]{F1F1F1} \raisebox{-1pt}{\small 0.68}} & \makebox[\linewidth]{{\cellcolor[HTML]{4F81BA}} \color[HTML]{F1F1F1} \raisebox{-1pt}{\small 0.54}} \\
16 & \makebox[\linewidth]{{\cellcolor[HTML]{78B0D3}} \color[HTML]{F1F1F1} \raisebox{-1pt}{\small 0.72}} & \makebox[\linewidth]{{\cellcolor[HTML]{426CB0}} \color[HTML]{F1F1F1} \raisebox{-1pt}{\small 0.48}} & \makebox[\linewidth]{{\cellcolor[HTML]{3D5BA7}} \color[HTML]{F1F1F1} \raisebox{-1pt}{\small 0.44}} & \makebox[\linewidth]{{\cellcolor[HTML]{C7E7F1}} \color[HTML]{000000} \raisebox{-1pt}{\small 1.20}} \\
32 & \makebox[\linewidth]{{\cellcolor[HTML]{578ABF}} \color[HTML]{F1F1F1} \raisebox{-1pt}{\small 0.57}} & \makebox[\linewidth]{{\cellcolor[HTML]{8ABEDA}} \color[HTML]{000000} \raisebox{-1pt}{\small 0.81}} & \makebox[\linewidth]{{\cellcolor[HTML]{8ABEDA}} \color[HTML]{000000} \raisebox{-1pt}{\small 0.81}} & \makebox[\linewidth]{{\cellcolor[HTML]{DCF1F7}} \color[HTML]{000000} \raisebox{-1pt}{\small 1.36}} \\
64 & \makebox[\linewidth]{{\cellcolor[HTML]{6399C7}} \color[HTML]{F1F1F1} \raisebox{-1pt}{\small 0.62}} & \makebox[\linewidth]{{\cellcolor[HTML]{8EC2DC}} \color[HTML]{000000} \raisebox{-1pt}{\small 0.84}} & \makebox[\linewidth]{{\cellcolor[HTML]{FFF3AD}} \color[HTML]{000000} \raisebox{-1pt}{\small 2.26}} & \makebox[\linewidth]{{\cellcolor[HTML]{FEDA8A}} \color[HTML]{000000} \raisebox{-1pt}{\small 2.92}} \\
128 & \makebox[\linewidth]{{\cellcolor[HTML]{AAD8E9}} \color[HTML]{000000} \raisebox{-1pt}{\small 1.00}} & \makebox[\linewidth]{{\cellcolor[HTML]{FEDA8A}} \color[HTML]{000000} \raisebox{-1pt}{\small 2.90}} & \makebox[\linewidth]{{\cellcolor[HTML]{FDB366}} \color[HTML]{000000} \raisebox{-1pt}{\small 3.83}} &  \\
\bottomrule
\end{tabularx}
\captionsetup{skip=2pt}
\caption{N = 3}
\end{subtable}
\hfill
\begin{subtable}[h]{0.05\linewidth}
\centering
\includegraphics[height=70pt]{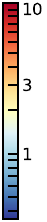}
\vfill \vspace{-0.5em}
\end{subtable}
\hfill
\caption{Median values of the training time (in hours) on the Threshold-Ratio dataset.}
\label{tab:time_median_threshratio}
\vspace{-0.5em}\end{table*}